\documentclass[11pt]{article}

\usepackage{fullpage,parskip}
\usepackage[utf8]{inputenc} 
\usepackage[T1]{fontenc}    
\usepackage[hidelinks,colorlinks=true,citecolor=black!30!blue,linkcolor=black!10!red]{hyperref}       
\usepackage{url}            
\usepackage{enumitem}
\usepackage{booktabs}       
\usepackage{amsfonts}       
\usepackage{nicefrac}       
\usepackage{microtype}      
\usepackage{xcolor}         

\usepackage{hyperref}
\usepackage{graphicx}

\usepackage{wrapfig}
\usepackage[utf8]{inputenc} 
\usepackage[T1]{fontenc}    
\usepackage{hyperref,xcolor}       
\usepackage{url}            
\usepackage{booktabs}       
\usepackage{amsfonts}       
\usepackage{nicefrac}       
\usepackage{microtype}      
\usepackage{xcolor}         
\usepackage{comment}

\usepackage{amsmath,amsfonts,amssymb}
\usepackage{graphicx}

\usepackage{algorithm}
\usepackage{algorithmic}

\usepackage{multirow}

\usepackage{bm}
\usepackage[all]{hypcap}
\usepackage{caption}
\usepackage{subcaption}
\usepackage{mathtools} 
\usepackage{mleftright}  

\usepackage{amsthm, thmtools}
\usepackage{amsmath, amsfonts, mathtools, amssymb}
\usepackage{float}
\newtheoremstyle{sltheorem}
                {}          
                {}          
                {\slshape}  
                {}          
                {\bfseries} 
                {.}         
                { }         
                {}          
\theoremstyle{sltheorem}

\DeclareMathOperator*{\argmin}{arg\,min}

\usepackage{xspace}
\usepackage{bbm}

\def\diag{\mathop{\rm diag}\nolimits}%
\def\Re{\mathop{\rm Re}\nolimits}%

\newcommand{\Xb}{\mathbf{X}}

\newcommand{\wv}{{\bf w}}

\newcommand{\Ac}{\mathcal{A}}

\newcommand{\Nc}{\mathcal{N}}

\newcommand{\Xc}{\mathcal{X}}
\newcommand{\Lc}{\mathcal{L}}

\newcommand{\bv}{{\bf b}}
\newcommand{\gv}{{\bf g}}
\newcommand{\hv}{{\bf h}}

\newcommand{\xv}{{\bf x}}
\newcommand{\mv}{{\bf m}}
\newcommand{\yv}{{\bf y}}
\newcommand{\zv}{{\bf z}}
\newcommand{\uv}{{\bf u}}
\newcommand{\vv}{{\bf v}}
\newcommand{\sv}{{\bf s}}
\newcommand{\dv}{{\bf d}}

\DeclareMathOperator\E{E}

\def\textiid{i.i.d.\@\xspace}
\newcommand\iid{\ifmmode\text{ i.i.d. } \else \textiid \fi}

\title{Monte Carlo Maximum Likelihood Reconstruction for Digital Holography with Speckle}

\author{Xi Chen, Arian Maleki, Shirin Jalali}

\date{}

\begin{document}

\maketitle

\renewcommand\thefootnote{}
\footnotetext{
Xi Chen and Shirin Jalali are with the Department of Electrical and Computer Engineering,
Rutgers University, Piscataway, NJ 08854 USA
(e-mail: xi.chen15@rutgers.edu; shirin.jalali@rutgers.edu).}

\footnotetext{
Arian Maleki is with the Department of Statistics,
Columbia University, New York, NY 10027 USA
(e-mail: arian@stat.columbia.edu).
}

\footnotetext{
Part of this work was presented at the 2025 IEEE 22nd International Symposium
on Biomedical Imaging (ISBI), Houston, TX.~\cite{chen2025monte}.
}

\begin{abstract}
In coherent imaging, speckle is statistically modeled as multiplicative noise, posing a fundamental challenge for image reconstruction. While maximum likelihood estimation (MLE) provides a principled framework for speckle mitigation, its application to coherent imaging system such as digital holography with finite apertures is hindered by the prohibitive cost of high-dimensional matrix inversion, especially at high resolutions. This computational burden has prevented the use of MLE-based reconstruction with physically accurate aperture modeling. In this work, we propose a randomized linear algebra approach that enables scalable MLE optimization without explicit matrix inversions in gradient computation. By exploiting the structural properties of sensing matrix and using conjugate gradient for likelihood gradient evaluation, the proposed algorithm supports accurate aperture modeling without the simplifying assumptions commonly imposed for tractability. We term the resulting method projected gradient descent with Monte Carlo estimation (PGD-MC). The proposed PGD-MC framework (i) demonstrates robustness to diverse and physically accurate aperture models, (ii) achieves substantial improvements in reconstruction quality and computational efficiency, and (iii) scales effectively to high-resolution digital holography. Extensive experiments incorporating three representative denoisers as regularization show that PGD-MC provides a flexible and effective MLE-based reconstruction framework for digital holography with finite apertures, consistently outperforming prior Plug-and-Play model-based iterative reconstruction methods in both accuracy and speed. Our code is available at: \url{https://github.com/Computational-Imaging-RU/MC_Maximum_Likelihood_Digital_Holography_Speckle}.
\end{abstract}

\section{Introduction}
\label{sec:intro}
Coherent imaging systems, such as synthetic aperture radar (SAR) ~\cite{argenti2013tutorial}, digital holography~\cite{bianco2018strategies}, optical coherence tomography (OCT) and ultrasound, are known to suffer from speckle, which is statistically modeled as multiplicative noise \cite{goodman2007speckle}. Speckle arises because coherent illumination interacts with object surfaces that are rough at the scales comparable to the illumination wavelength. The reflected wavefronts from these numerous scatterers interfere  when recorded by the sensor, producing the granular intensity variations known as speckle. Existence of speckle noise considerably degrades the images acquired by such systems and prevents them from generating high-quality images \cite{goodman2007speckle}.

Mathematically, the acquisition  process of a coherent imaging system in the presence of speckle can be modeled as
\begin{align} \label{eq:forward_single}
    \yv = A \mathbf{g} + \zv,
\end{align}
where $\yv \in \mathbb{C}^m$ is the measurement vector, $A \in \mathbb{C}^{m \times n}$ is the sensing matrix, $\zv \sim \mathcal{CN}(\mathbf{0},\sigma_z^2 I_m)$ is the additive sensor noise. $\mathbf{g} \sim \mathcal{CN}(\mathbf{0},X) \in \mathbb{C}^{n}$ is the complex reflectance image with speckle, where $X = \diag(\xv)$ denotes the diagonal matrix formed from the underlying speckle-free reflectivity $\xv$. Solving this inverse problem entails  reconstructing signal $\xv$ from measurements $\yv$, while knowing the sensing matrix $A$.

While \eqref{eq:forward_single} provides a statistically accurate description of the system, directly solving this inverse problem leads to a log-likelihood function  involving the determinant and inverse of $AXA^H$. Optimizing this objective is therefore computationally prohibitive. Consequently, most classical approaches avoid this formulation and instead rely on sub-optimal approximations.

A key class of standard  methods to address speckle noise relies on applying optical and numerical despeckling methods \cite{bianco2018strategies}. Optical solutions typically involve  multi-look acquisition for speckle reduction. For example, rotating a ground-glass diffuser introduces random spatial phase modulation, enabling the recording of  multiple looks under independent realizations of the speckle noise \cite{memmolo2014encoding}. For amplitude reconstruction, a variety of adaptive spatial filters have been proposed \cite{lee1980digital,frost1982model,kuan1985adaptive,uzan2013speckle}. More recently, end-to-end deep learning models are also developed  for despeckling \cite{zhang2018learning, joo2019dopamine} by utilizing supervised learning approach with paired training data. 

In coherent imaging systems such as digital holography, early works including \cite{sotthivirat2004penalized, lee2016single} focus on reconstructing a complex-valued object field directly from real-valued hologram measurements. These approaches primarily address phase retrieval and wavefront reconstruction and do not explicitly model speckle as a multiplicative noise process. In contrast, recent works \cite{pellizzari2017phase,pellizzari2019imaging,pellizzari2020coherent,bate2021model,allen2025clamp} seek to estimate the underlying speckle-free real-valued reflectivity by explicitly modeling speckle noise as in \eqref{eq:forward_single}. Learning-based image priors, such as DnCNN \cite{zhang2017beyond}, are incorporated  through the Plug-and-Play (PnP) framework \cite{venkatakrishnan2013plug} for model-based reconstruction. However, since the maximum likelihood estimation (MLE) of the underlying image given the measurement involves an intractable inversion of the covariance matrix $AXA^H$, the authors derived a variational lower bound of the likelihood function for optimization. In maximizing the variational lower bound, instead, only the computation of $A^HA$ is needed. An additional approximation $A^HA\approx I$ was therefore introduced to simplify the EM updates \cite{dempster1977maximum}. The resulting approximation errors in both the likelihood function and the sensing matrix can lead to sub-optimal reconstruction performance. Moreover, the coordinate-wise updates used in the M-step hinder scalability to high-resolution imaging systems.

More recently, \cite{zhou2022compressed,chenbagged,chen2025multilook} have theoretically and algorithmically demonstrated that directly solving the MLE optimization, even if forward model is ill-posed ($m<n$), is feasible.  To mitigate the computational complexity of the  matrix inversion involved in the MLE objective, \cite{chenbagged} employed the Newton-Schulz iterative algorithm \cite{gower2017randomized,stotsky2020efficient}. This approach enabled, for the first time, direct optimization of the MLE objective. However, the proposed algorithm requires forming and manipulating matrices such as $AXA^H$, whose large dimensionality limits the applicability to high-resolution imaging scenarios.

\begin{figure}[t]
    \centering

    {\footnotesize
    \subfloat[Circular aperture]{
        \includegraphics[width=0.38\columnwidth]{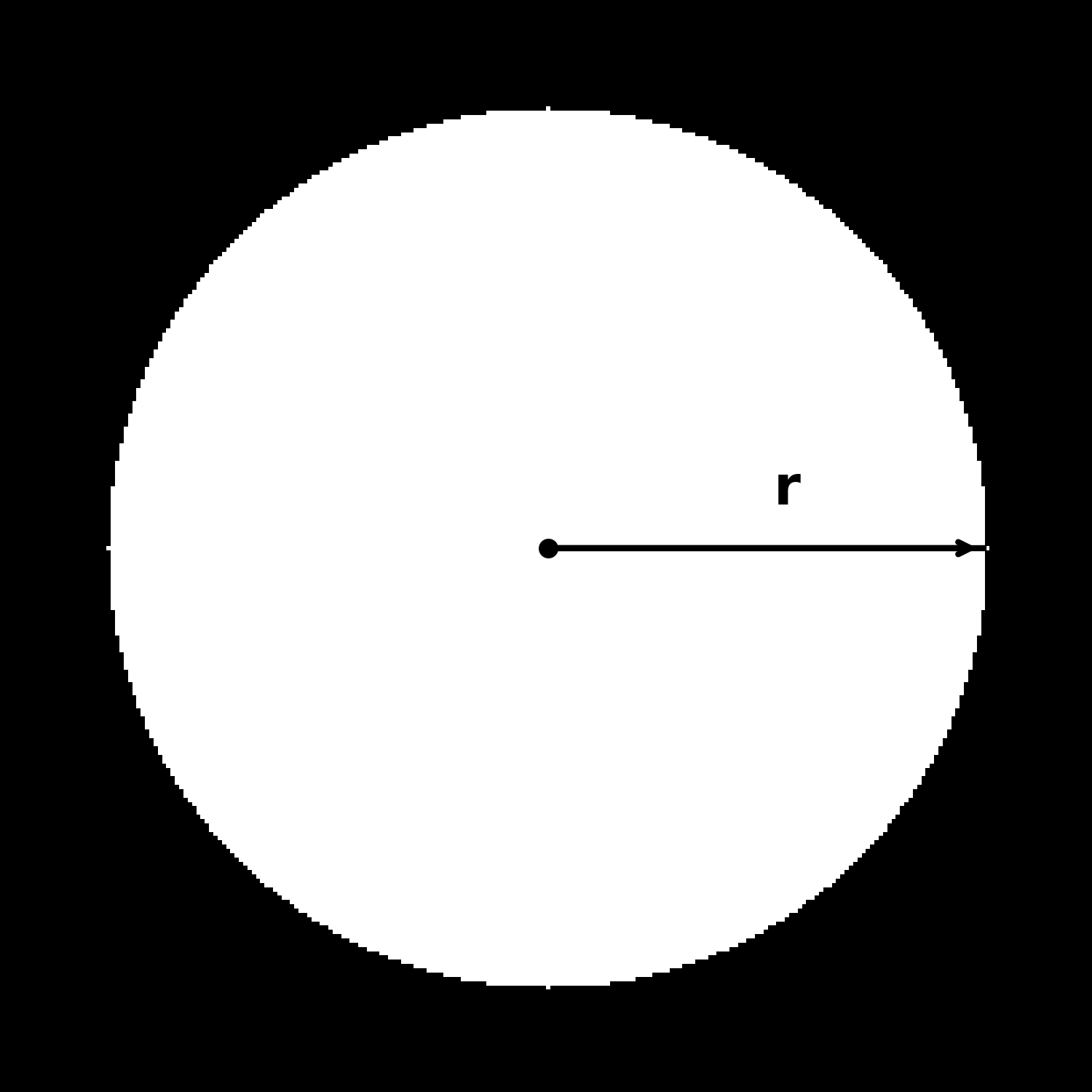}
    }
    \hspace{-0.4em}
    \subfloat[Annular aperture]{
        \includegraphics[width=0.38\columnwidth]{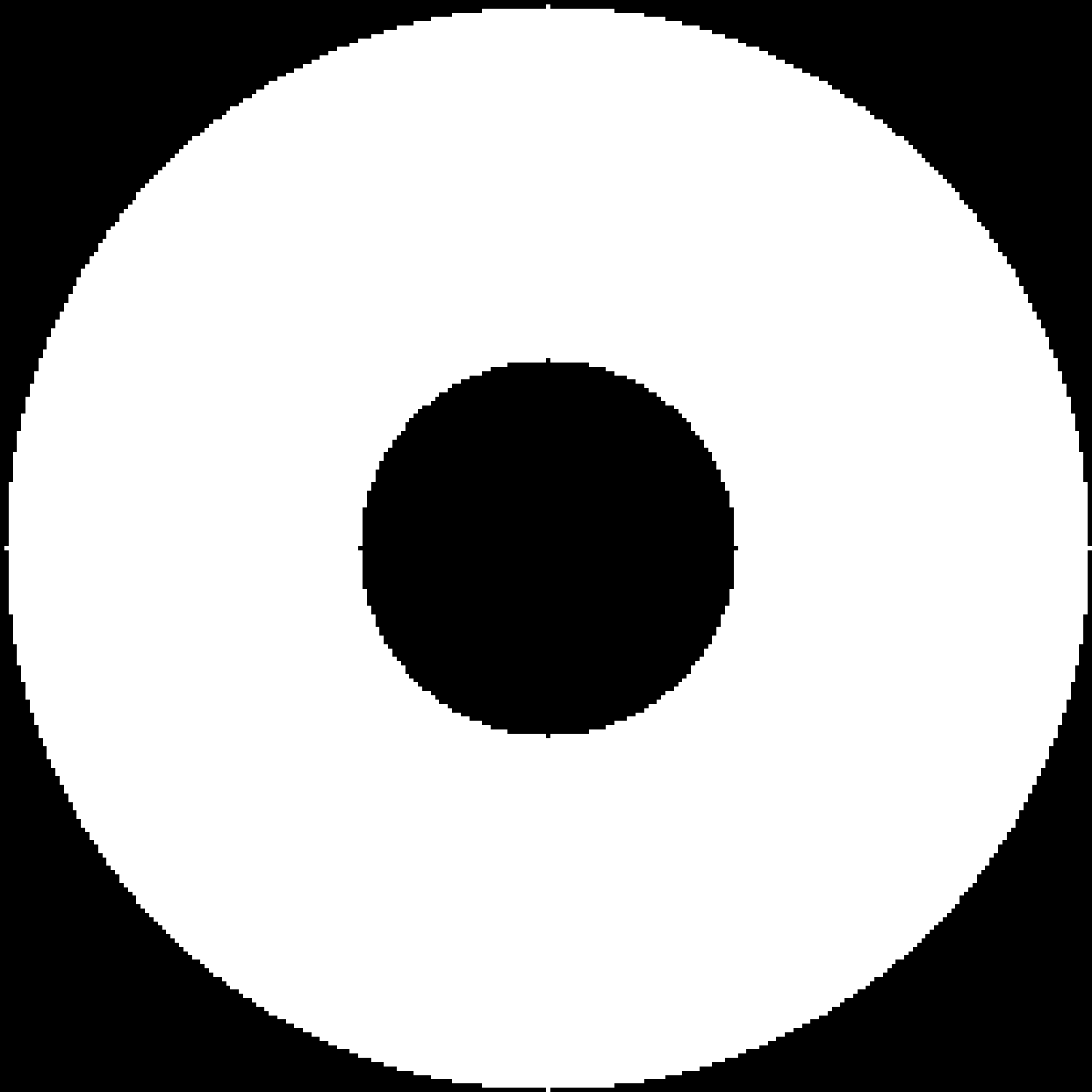}
    }
    }

    \caption{Visualization of apertures in the imaging forward model \eqref{eq:forward_1}.}
    \label{fig:apertures}
\end{figure}

In this work, we focus on digital holography as a key application of coherent imaging and develop computationally efficient methods for directly solving the MLE-based reconstruction problem under a physically accurate speckle model. Specifically, we consider the problem of reconstructing a speckle-free, real-valued reflectivity image from complex-valued holographic measurements acquired with practical circular and annular apertures. (See Fig.~\ref{fig:apertures}.) We formulate a constrained MLE problem and solve it using projected gradient descent (PGD). To overcome the prohibitive matrix inversion required by the MLE objective, we propose a randomized linear algebra strategy to compute the gradient of the log-likelihood using Monte Carlo-based trace estimation and conjugate gradient method \cite{golub2013matrix,kireeva2024randomized}. This approach avoids explicit formation and inversion of $AXA^H$ without making any simplifying assumptions.

To demonstrate the flexibility of the proposed framework, we incorporate three representative image denoisers as regularization within the PGD scheme. Furthermore, by exploiting the Fourier structure of the holographic forward operator, we implement operations using fast Fourier transforms, avoiding explicit vectorization and matrix–vector multiplications. As a result, the proposed projected gradient descent with Monte Carlo estimation (PGD-MC) scales efficiently to high-resolution settings, and we demonstrate direct MLE-based reconstruction for digital holography images with resolutions up to $512 \times 512$.

\section{Problem formulation}
\label{sec:background}

\subsection{Forward model in coherent imaging with speckle}

We consider a multi-look coherent imaging system where multiple sets of measurements are acquired. Let $X \in (\mathbb{R}^+)^{H \times W}$ denote the desired 2D image, where $H$ and $W$ denote the height and width of it,  respectively. Let  $\xv ={\rm Vec}(X) \in (\mathbb{R}^+)^{n}$  denote its vectorized form, where   $n = HW$. The multi-look measurement model with independent speckle realizations is given by
\begin{align} \label{eq:forward_1}
    \yv_\ell = A \gv_\ell + \zv_\ell, \quad \ell=1,\cdots,L,
\end{align}
where $\yv_\ell \in \mathbb{C}^m$ is the $\ell$-th measurement vector, $A \in \mathbb{C}^{m \times n}$ is the constant sensing matrix across looks, $\zv_\ell \sim \mathcal{CN}(\mathbf{0},\sigma_z^2 I_m)$ is the additive sensor noise. $\mathbf{g}_\ell \sim \mathcal{CN}(\mathbf{0},X) \in \mathbb{C}^{n}$ is complex reflectance image with independent speckle realization. Under this model, the expected value of the square magnitude of $\gv_{\ell}$ is the intensity of the image, i.e., $\mathbb{E}[|\gv_{\ell}|^2] = \xv$. The problem is to estimate $\xv$, from measurements $\yv_1,\ldots,\yv_L$, while  knowing the forward operator $A$.

A key application of \eqref{eq:forward_1} is digital holography. In this setting, the number of measurements equals the number of unknowns, i.e., $m = n$, and the forward operator can be expressed as
\begin{align} \label{eq:sensing_matrix}
    A = \mathcal{F}^{-1} M \mathcal{F},
\end{align}
where $\mathcal{F} \in \mathbb{C}^{n \times n}$ denotes the discrete Fourier transform operator \cite{goodman2005introduction}. In practice, the Fourier transform is applied to the 2D image. When acting on the vectorized image $\xv$, the operator $\mathcal{F}$ can be written as the Kronecker product of row and column Fourier matrices, i.e.,
\[
\mathcal{F} = \mathcal{F}_r \otimes \mathcal{F}_c^H.
\]
The matrix $M = \mathrm{diag}(\mathrm{vec}(P)) \in \{0,1\}^{n \times n}$ represents the aperture mask in the spatial frequency domain, where $P \in \{0,1\}^{H \times W}$ denotes the 2D aperture function. In the classic setup, $P$ corresponds to a circular aperture that acts as a low-pass filter in the spatial frequency domain and is defined as
\[
P(h,w) =
\begin{cases}
1, & \sqrt{(h-h_0)^2 + (w-w_0)^2} \le r,\\[4pt]
0, & \text{otherwise},
\end{cases}
\]
where $(h_0, w_0)$ denotes the aperture center and $r$ is the radius. To improve generality and account for central obscuration encountered in practical optical systems, we also consider annular apertures \cite{goodman2005introduction}. The aperture configurations used in this work are illustrated in Fig.~\ref{fig:apertures}. Although $A$ is square, the presence of the aperture mask generally renders it  rank-deficient, and the resulting inverse problem remains ill-posed.

\subsection{Maximum likelihood estimation for image reconstruction}

Given the independence and Gaussianity of  $\zv_{\ell}$ and $\gv_{\ell}$, with fixed $\xv$ and $A$, the measurement vectors $\yv_1,\ldots,\yv_L$ are independently distributed as zero-mean Gaussian random vectors with  covariance  $\Sigma(\xv)$ defined as 
\begin{gather}
     \Sigma(\xv)= AXA^H + \sigma_z^2 I_m.
\end{gather}
Given $\xv$, the  negative log-likelihood of  $\yv_\ell$, $ -\log p(\yv_\ell|A,X)$,  up to  additive constants independent of $\xv$, is denoted as $f_\ell(\xv)$, which can be written as
\[
f_\ell(\xv)  = \log \text{det} \Sigma(\xv) + \yv_\ell^H \Sigma^{-1}(\xv) \yv_\ell,
\]
where $\yv_\ell^H$ is the Hermitian transpose of $\yv_\ell$. Given the independence of the measurements, the normalized negative log-likelihood of all  $L$ looks can be   written as $f_L(\xv)={\frac{1}{L}}\sum_{\ell}f_{\ell}(\xv)$. That is,
\begin{align} \label{eq:objective}
f_L(\xv) = \log \text{det} \Sigma(\xv) + \frac{1}{L}\sum^L_{\ell=1} \yv_\ell^H \Sigma^{-1}(\xv) \yv_\ell.
\end{align}
It is well known that properly modeling and exploiting the underlying source structure plays a key role in solving noisy and ill-posed inverse problems, and can substantially improve reconstruction performance. Motivated by this, instead of optimizing over all $\xv \in (\mathbb{R}^+)^n$, we assume that $\xv$ belongs to a class of images $\Xc$ that captures prior knowledge of the desired signal. This leads to the constrained MLE formulation
\begin{align}
\label{eq:MLE_constrained}
    \hat{\xv} = \argmin_{\xv \in \Xc} f_L(\xv).
\end{align}
The choice of $\Xc$ plays a central role in regularizing the reconstruction and mitigating the effects of speckle noise and limited measurements.

A flexible and practically effective way to impose such structural constraints is through image denoisers, which implicitly define the set $\Xc$ of admissible images \cite{romano2017little}. Both classical non-learning-based denoisers, such as BM3D \cite{dabov2007image}, and learning-based methods, including pre-trained DnCNN \cite{zhang2017beyond} and untrained neural networks (UNNs) such as deep image prior (DIP) \cite{ulyanov2018deep,heckel2018deep}, have been widely used as regularizers in inverse problems. These approaches differ in how $\Xc$ is represented and enforced, but share the common goal of restricting the solution space to images with desirable structure. 

In the present work, we develop an optimization framework that accommodates multiple choices of structural priors, including UNN-based parameterizations as well as plug-and-play denoisers such as BM3D and DnCNN. This generality allows us to systematically study the impact of different source models within a unified MLE-based reconstruction framework.  We next describe the proposed algorithm for efficiently solving \eqref{eq:MLE_constrained}.
\subsection{Projected gradient descent to solve the optimization problem}

To solve the nonconvex constrained MLE problem in \eqref{eq:MLE_constrained}, we adopt the PGD approach. Starting from an initial estimate $\xv_0$, the iterations are given by
\begin{align}
    \sv_{t+1} &= \xv_t - \mu \nabla f_L(\xv_t), \label{eq:pgd_grad}\\
    \xv_{t+1} &= \Pi_{\Xc}(\sv_{t+1}), \label{eq:pgd_proj}
\end{align}
where $\mu>0$ denotes the step size, $\nabla f_L(\xv_t)$ is the gradient of the negative log-likelihood at iteration $t$, and $\Pi_{\Xc}(\cdot)$ denotes a projection onto the constraint set $\Xc$. The gradient descent step enforces data fidelity through the likelihood function, while the projection step incorporates prior information by restricting the iterate to the admissible set $\Xc$. The gradient of $f_L$ with respect to $\xv_t$ is given by (Refer to Appendix \ref{sec:app_A})
\begin{align} \label{eq:grad}
   \nabla f_L(\xv_t)
   = \diag\!\left(A^H \Sigma(\xv_t)^{-1} A \right)
   - \frac{1}{L}\sum^L_{\ell=1} \left|A^H \Sigma(\xv_t)^{-1} \yv_\ell \right|^{2},
\end{align}
where the absolute value and square are taken elementwise.


\subsection{Challenges in MLE-based reconstruction}

The primary computational bottleneck in evaluating $\nabla f_L(\xv_t)$ is the inversion of the $m \times m$ covariance matrix $\Sigma(\xv_t)$. Existing approaches address this issue either by optimizing a variational lower bound of the likelihood, relying on simplifying assumptions such as $A^H A \approx I$ \cite{pellizzari2020coherent}, or by directly approximating $\Sigma(\xv_t)^{-1}$ using iterative matrix inversion methods such as Newton-Schulz \cite{chenbagged}. While the former leads to surrogate objectives and potential model mismatch, the latter requires storage of large matrices, which limits scalability.


In this work, we exploit the structure of the forward operator $A$ in \eqref{eq:sensing_matrix} and propose a randomized linear algebra approach, combined with the conjugate gradient method, to efficiently compute the gradient of the exact likelihood function. Moreover, once the likelihood gradient can be computed efficiently, the proposed framework naturally accommodates a wide range of prior models and regularization strategies through gradient-based optimization. Although we focus on PGD in this work, the proposed gradient computation is applicable to other first-order optimization methods for MLE-based reconstruction in coherent imaging with speckle.


\section{Randomized matrix-Free gradient evaluation for MLE}
In this section, we introduce the proposed approach to efficiently compute the gradient of the likelihood function. The gradient of $f_L(\xv)$ defined in \eqref{eq:grad} involves two terms: the diagonal entries of $A^H\Sigma(\xv_t)^{-1}A$, and $A^H \Sigma(\xv_t)^{-1} \yv_\ell$. We first introduce how to use conjugate gradient method to calculate the second term. Then we introduce how to use randomized linear algebra to approximate the diagonal terms of the first term, and further use conjugate gradient to avoid matrix inversion.

\subsection{Conjugate gradient}
To compute $A^H \Sigma(\xv_t)^{-1} \yv_\ell$, $\ell=1,\cdots,L$ in \eqref{eq:grad}, define  $\hv^{(\ell)}_{t}=\Sigma(\xv_t)^{-1} \yv_\ell$, $\ell=1,\ldots,L$. Note that $\hv^{(\ell)}_{t}$ is the solution of the following  least-squares optimization:
\begin{align}\label{eq:cg_1}
    \operatorname*{argmin}_{\mathbf{h}} \| \Sigma(\xv_t) \mathbf{h} - \yv_\ell \|^2.
\end{align} 
Therefore, if we are able to efficiently solve \eqref{eq:cg_1}, we can get the desired term $A^H \Sigma(\xv_t)^{-1} \yv_\ell$ by just applying $A^H$ to  $\hat{\mathbf{h}}^{(\ell)}_{t}$.
To efficiently solve \eqref{eq:cg_1}, we use the conjugate gradient method \cite{golub2013matrix}, which leverages the fact that $\Sigma(\xv_t)$ is Hermitian positive definite. The details are outlined in Algorithm \ref{alg:CG}.

\begin{algorithm}[t]
\caption{Conjugate gradient for 
$\operatorname*{argmin}_{\mathbf{h}} 
\| \Sigma(\xv_t)\mathbf{h} - \yv_\ell \|^2$}
\label{alg:CG}
\begin{algorithmic}[1]

\STATE {\bfseries Initialize} 
$\mathbf{h}_0$, $\mathbf{r}_0 = \yv_\ell - \Sigma(\xv_t)\mathbf{h}_0$,
initial direction $\mathbf{p}_0 = \mathbf{r}_0$,
stopping tolerance $\epsilon$.

\FOR{$t = 0,1,\ldots,T-1$}
    \STATE $
    \alpha_t =
    \frac{\mathbf{r}_t^{H}\mathbf{r}_t}
         {\mathbf{p}_t^{H}\Sigma(\xv_t)\mathbf{p}_t}
    $
    \hfill (step size)
    \STATE $
    \mathbf{h}_{t+1} = \mathbf{h}_t + \alpha_t \mathbf{p}_t
    $
    \hfill (solution update)
    \STATE $
    \mathbf{r}_{t+1} = \mathbf{r}_t - \alpha_t \Sigma(\xv_t)\mathbf{p}_t
    $
    \hfill (residual update)
    \IF{$\|\mathbf{r}_{t+1}\|_2 \le \epsilon$}
        \STATE $\hat{\mathbf{h}} = \mathbf{h}_{t+1}$ \textbf{break}
        \hfill (convergence)
    \ENDIF
    \STATE $
    \beta_t =
    \frac{\mathbf{r}_{t+1}^{H}\mathbf{r}_{t+1}}
         {\mathbf{r}_t^{H}\mathbf{r}_t}
    $
    \hfill (direction weight)
    \STATE $
    \mathbf{p}_{t+1} = \mathbf{r}_{t+1} + \beta_t \mathbf{p}_t
    $
    \hfill (new search direction)
\ENDFOR
\STATE {\bfseries Output} $\hat{\mathbf{h}} = \mathbf{h}_T$
\end{algorithmic}
\end{algorithm}

Running Algorithm~\ref{alg:CG} requires the ability to compute expressions of the form $A\xv$ and $\Sigma(\xv)\hv$. Owing to the special structure of the forward operator, both operations can be carried out with low computational cost. Specifically, given a 2D image $\Xb$ and its vectorized form $\xv = \operatorname{vec}(\Xb)$,
\[
A \xv = \operatorname{vec}(\Ac(\Xb)), \; \text{where} \; 
\Ac(\Xb) = \operatorname{IFFT2}\!\left(P \odot \operatorname{FFT2}(\Xb)\right).
\]
Here, $\operatorname{FFT2}$ denotes the two-dimensional Fourier transform, and $P$ represents the 2D aperture matrix defined earlier.
Similarly, the matrix--vector product $\Sigma(\xv)\hv$ is equivalent to applying the  operator of $\Sigma(\xv)$ to the 2D version of $\hv$, denoted by $\mathbf{H}$, as
\[
\Sigma(\xv)\hv
= \operatorname{vec}\!\left(\Ac\!\left(\Xb \odot \Ac(\mathbf{H})\right) + \sigma_z^2 \mathbf{H}\right).
\]
Here, $\Xb$ denotes the 2D (image) representation of $\xv$. Note that by operating directly in the 2D domain, computations can be performed efficiently without explicit vectorization.  Exploiting the properties of the Fourier transform in \eqref{eq:sensing_matrix} and using fast Fourier transform algorithms within the conjugate gradient method accelerate  reconstruction and improve  scalability to larger images.

\begin{algorithm}[t]
\caption{PGD-MC with conjugate gradient for MLE-based coherent imaging reconstruction}
\label{alg:PGD}
\begin{algorithmic}[1]
\STATE {\bfseries Initialize} 
$\xv_{0}=\frac{1}{L} \sum^L_{\ell=1}|A^H \yv_\ell|^2$, 
projection $\Pi_{\Xc}(\cdot)$.
\FOR{$t=0,1,\ldots,T-1$}
    \STATE {\bfseries [Gradient descent step]}
    \STATE $\Sigma(\xv_t) = A X_t A^H + \sigma_z^2 I_m$
    \STATE {\bfseries Monte-Carlo approximation}
   \FOR{$k=1,2,\ldots,K$}
    \STATE $\vv^{(k)}_t \sim \Nc(0,I_n)$
    \STATE $\hat{\uv}^{(k)}_t \leftarrow 
    \operatorname*{argmin}_{\uv} 
    \| \Sigma(\xv_t)\uv - A\vv^{(k)}_t \|^2$
    \hfill (Apply CG)
    \STATE $\wv^{(k)}_t \leftarrow A^H\hat{\uv}^{(k)}_t$
\ENDFOR
\STATE $\dv_t \leftarrow \frac{1}{K}\sum_{k=1}^K \wv^{(k)}_t \odot \vv^{(k)}_t$

\FOR{$\ell=1,2,\ldots,L$}
    \STATE $\hat{\mathbf{h}}^{(\ell)}_t \leftarrow 
    \operatorname*{argmin}_{\mathbf{h}} 
    \| \Sigma(\xv_t)\mathbf{h} - \yv_\ell \|^2$ \hfill (Apply CG)
    \STATE $\mv^{(\ell)}_t \leftarrow A^H\hat{\mathbf{h}}^{(\ell)}_t$
\ENDFOR

    \STATE $\widehat{\nabla f_L}(\xv_t)
 \leftarrow 
\dv_t - \frac{1}{L}\sum_{\ell=1}^L |\mv^{(\ell)}_t|^2$

    \STATE $\sv_{t+1} \leftarrow \xv_t - \mu\widehat{\nabla f_L}(\xv_t)
$ 
    \STATE {\bfseries [Projection step]}
    \STATE $\xv_{t+1} \leftarrow \Pi_{\Xc}(\sv_{t+1})$
\ENDFOR
\STATE {\bfseries Output} $\hat{\xv} = \xv_T$
\end{algorithmic}
\end{algorithm}



\begin{figure*}[t]
    \centering
    {\footnotesize

    \begin{subfigure}{0.39\textwidth}
        \centering
        \includegraphics[width=0.49\linewidth]{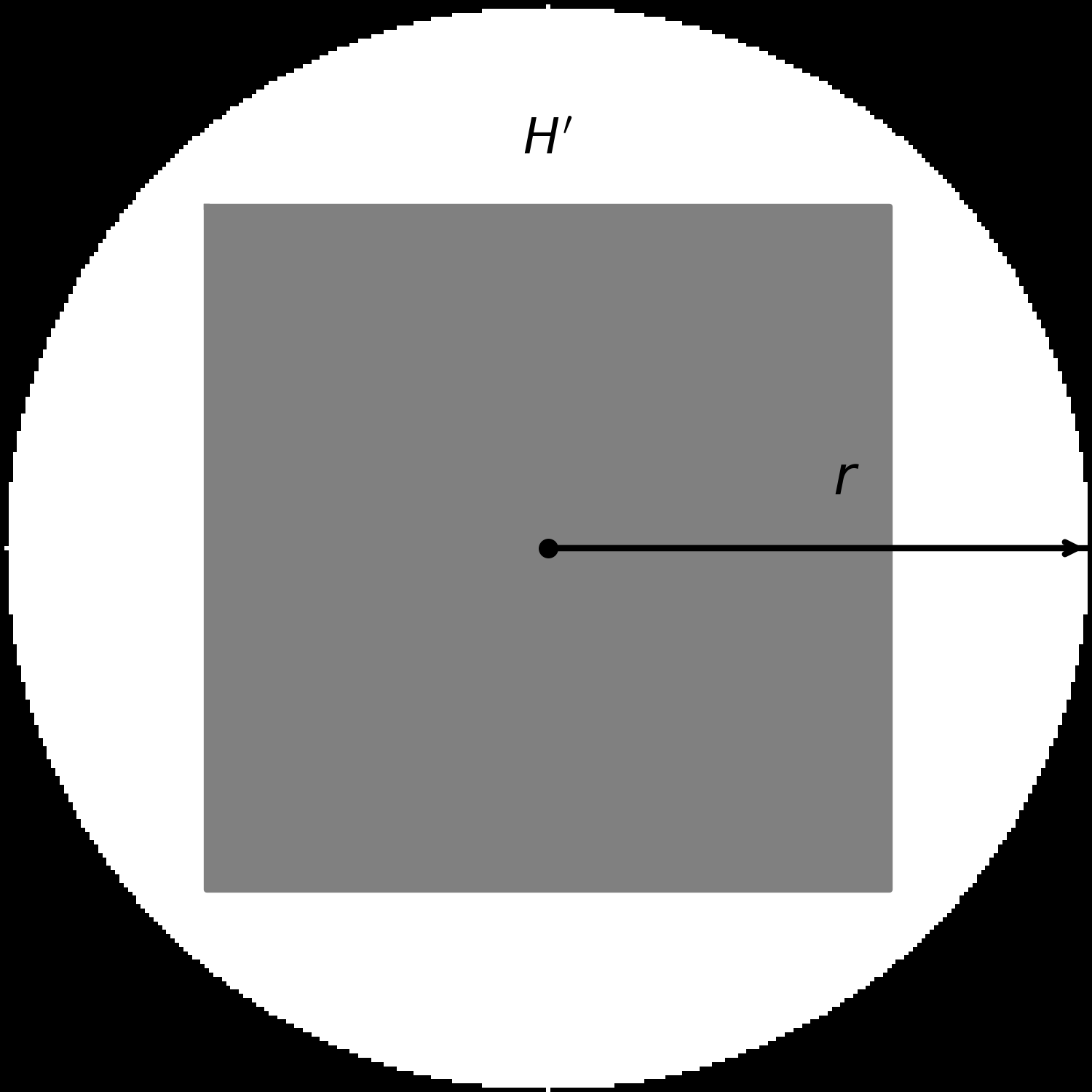}
        \hspace{-0.6em}
        \includegraphics[width=0.49\linewidth]{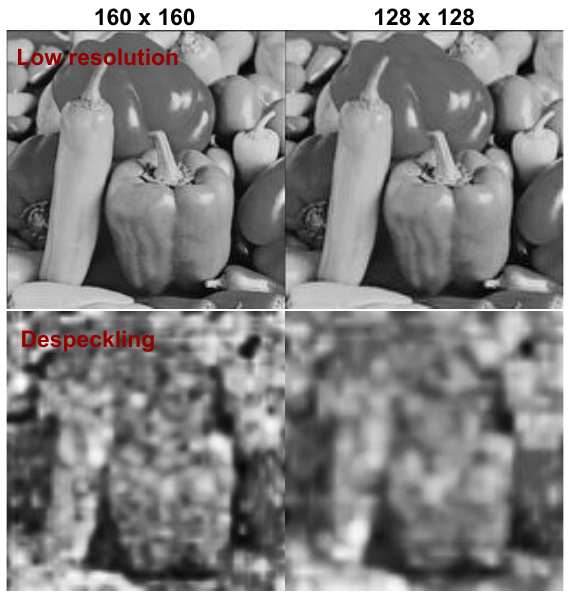}
        \caption{}
    \end{subfigure}
    \hspace{-0.6em}
    \begin{subfigure}{0.6\textwidth}
        \centering
        \includegraphics[width=0.33\linewidth]{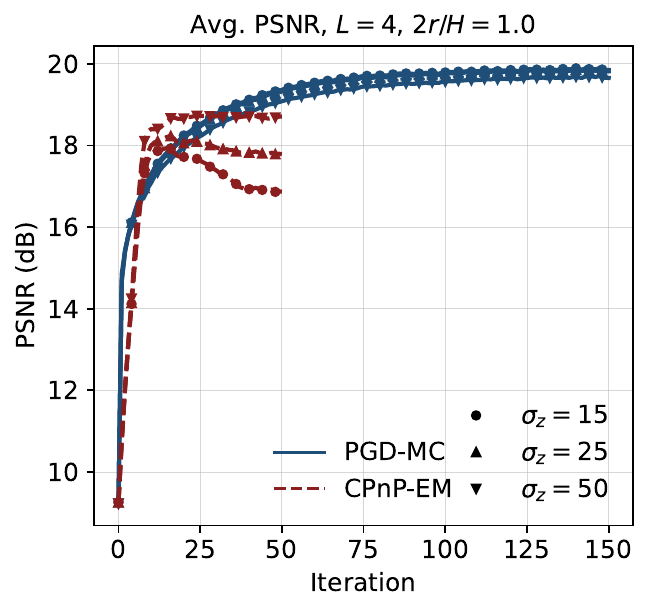}
        \hspace{-0.8em}
        \includegraphics[width=0.33\linewidth]{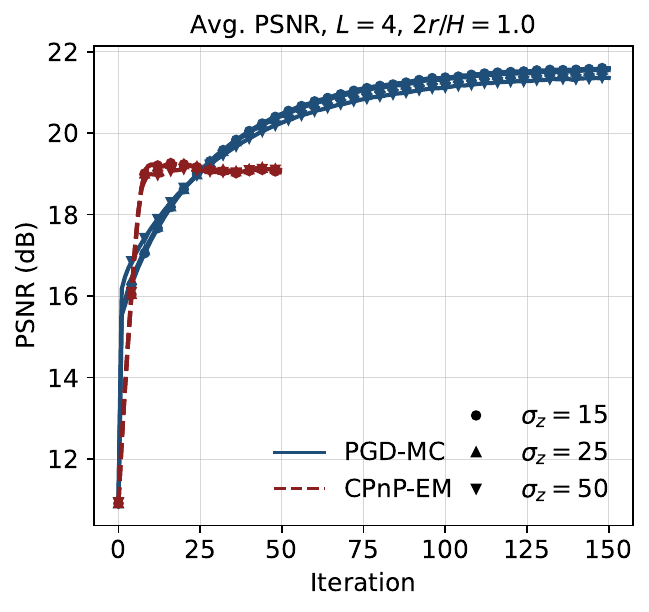}
        \hspace{-0.8em}
        \includegraphics[width=0.33\linewidth]{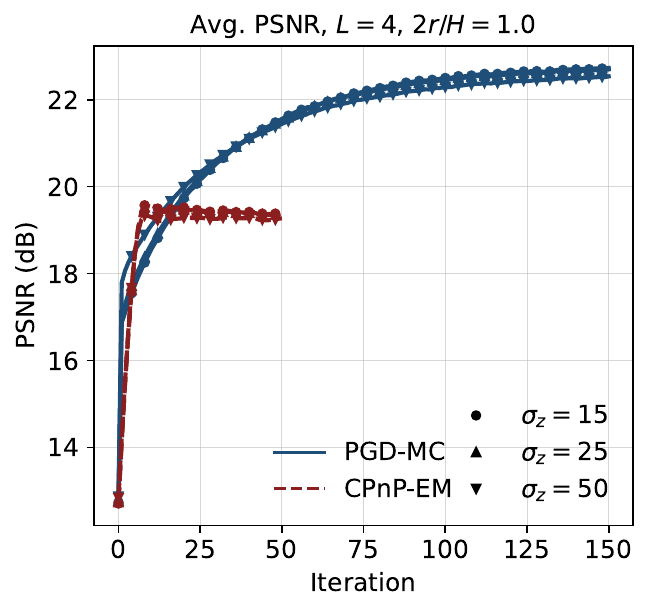}
        \caption{}
    \end{subfigure}
    }
    \caption{
        (a) Sub-area of hologram spectrum used for the corresponding low-resolution despeckling; (b) Average PSNR convergence curves of PGD-MC and CPnP-EM, with different noise levels $\sigma_z=15,25,50$, and numbers of looks $L=1,2,4$, where prior model is Deep Decoder.}
    \label{fig:subarea_psnr_convergence}
\end{figure*}

\subsection{Monte Carlo-based approximation of matrix diagonals}

Next, we propose a method based on randomized linear algebra to estimate the diagonal entries of $A^H \Sigma(\xv_t)^{-1} A$ without the explicit $\Sigma(\xv_t)$ or its inverse. To this end, let $\vv \sim \Nc(0, I_n)$ and define $\uv$ by $\uv = \Sigma(\xv_t)^{-1} A \vv.$
As before, $\uv$ can be computed as the solution to the following least-squares
\begin{align} \label{eq:cg_2}
    \hat{\uv} = \operatorname*{argmin}_{\uv} \bigl\| \Sigma(\xv_t)\uv - A\vv \bigr\|^2 .
\end{align}
Note that, since the entries of $\vv$ are zero-mean and independent,
\begin{align}
    \E\!\left[A^H \uv \odot \vv\right]
    = \E\!\left[(A^H \Sigma(\xv_t)^{-1} A \vv)\odot \vv\right]
\end{align}
is equal to the diagonal entries of $A^H \Sigma(\xv_t)^{-1} A$. This observation implies that $A^H \uv \odot \vv$ provides an unbiased estimator of the desired diagonal entries. 

To exploit this property and efficiently estimate the diagonal entries, we adopt a Monte Carlo approach. Specifically, we independently generate $\vv^{(1)},\ldots,\vv^{(K)}$, each distributed as $\mathcal{N}(\mathbf{0}, I_n)$, and compute $\uv^{(k)}$, for $k=1,\ldots,K$, by solving the corresponding least-squares problem. The desired diagonal entries are then estimated as
\begin{align} \label{eq:MC}
    \frac{1}{K} \sum_{k=1}^K A^H \hat{\uv}^{(k)}_{t} \odot \vv^{(k)}_{t}.
\end{align}
In practice, for sufficiently large $K$, this approximation yields an accurate estimate, as demonstrated later.

Algorithm~\ref{alg:PGD} summarizes the complete PGD framework incorporating the Monte Carlo approximation and conjugate gradient. This framework enables direct computation of the likelihood gradient without forming large matrices and performing expensive matrix–vector multiplications. It also avoids the memory overhead associated with storing such matrices. The resulting gradient descent step is efficient and accurate.

\section{Numerical Results}\label{sec:simulations}

\begin{table*}[t] 
\caption{Reconstruction performance comparison of CPnP-EM~\cite{pellizzari2020coherent} and PGD-MC (ours) on different aperture configurations, AWGN noise level $\sigma_z$, number of looks $L$, and prior models: BM3D, DnCNN, Deep Decoder. Average PSNR(dB) and SSIM are reported on seven test images. Best results are in \textbf{bold}, second-best are \underline{underlined}.} 
\begin{center}
\resizebox{\textwidth}{!}{
\begin{tabular}{cccccccccccc}
    \toprule
    \multicolumn{1}{c}{Aperture} & 
    \multicolumn{1}{c}{Noise level} & 
    \multicolumn{1}{c}{Looks} & 
    \multicolumn{1}{c}{Initialization} &
    \multicolumn{2}{c}{BM3D} &
    \multicolumn{2}{c}{DnCNN} &
    \multicolumn{2}{c}{Deep Decoder} \\
    \cmidrule(lr){1-1}
    \cmidrule(lr){2-2}
    \cmidrule(lr){3-3}
    \cmidrule(lr){4-4}
    \cmidrule(lr){5-6}
    \cmidrule(lr){7-8}
    \cmidrule(lr){9-10}

    Circular \& Annular & $\sigma_z$ & $L$ & $\frac{1}{L} \sum^L_{\ell=1}|A^H \yv_\ell|^2$ 
    & CPnP-EM & PGD-MC 
    & CPnP-EM & PGD-MC 
    & CPnP-EM & PGD-MC \\
    \midrule

\multirow{9}{*}{\shortstack{Circular $\frac{2r}{H}=0.8$\\ transparency rate $\approx 0.508$}}
    & \multirow{3}{*}{15}
        & 1 & 8.60 / 0.0789 & 10.29 / 0.2300 & \textbf{19.69} / \textbf{0.5726} 
          & 9.66 / 0.1215 & 11.28 / 0.4256 
          & 16.94 / 0.2242 & \underline{19.06} / \underline{0.4631} \\
    &   & 2 & 9.57 / 0.1189 & 10.80 / 0.3748 & \textbf{21.11} / \textbf{0.6343} 
          & 10.34 / 0.3509 & 11.65 / 0.4346 
          & 18.75 / 0.3723 & \underline{20.50} / \underline{0.5054} \\
    &   & 4 & 10.30 / 0.1722 & 10.93 / 0.4716 & \textbf{21.76} / \textbf{0.6488} 
          & 10.66 / 0.4222 & 11.76 / 0.4366 
          & 19.32 / 0.4755 & \underline{21.50} / \underline{0.5618} \\
    \cmidrule(l){2-10}

    & \multirow{3}{*}{25}
        & 1 & 8.62 / 0.0780 & 10.29 / 0.2338 & \textbf{19.65} / \textbf{0.5706} 
          & 9.68 / 0.1203 & 11.26 / 0.4249 
          & 17.29 / 0.2414 & \underline{19.01} / \underline{0.4631} \\
    &   & 2 & 9.60 / 0.1178 & 10.74 / 0.3823 & \textbf{21.01} / \textbf{0.6313} 
          & 10.37 / 0.3558 & 11.63 / 0.4344 
          & 18.83 / 0.4002 & \underline{20.43} / \underline{0.5035} \\
    &   & 4 & 10.36 / 0.1711 & 10.84 / 0.4696 & \textbf{21.68} / \textbf{0.6462} 
          & 10.68 / 0.2132 & 11.75 / 0.4367 
          & 19.29 / 0.4786 & \underline{21.44} / \underline{0.5605} \\
    \cmidrule(l){2-10}

    & \multirow{3}{*}{50}
        & 1 & 8.69 / 0.0736 & 10.12 / 0.3105 & \textbf{19.30} / \textbf{0.5591} 
          & 9.92 / 0.2937 & 11.24 / 0.4244 
          & 18.29 / 0.3857 & \underline{18.78} / \underline{0.4583} \\
    &   & 2 & 9.76 / 0.1122 & 10.51 / 0.1755 & \textbf{20.58} / \textbf{0.6195} 
          & 10.48 / 0.1674 & 11.65 / 0.4344 
          & 18.83 / 0.4544 & \underline{20.16} / \underline{0.4964} \\
    &   & 4 & 10.62 / 0.1651 & 10.97 / 0.2149 & \textbf{21.22} / \textbf{0.6323} 
          & 10.91 / 0.2065 & 11.78 / 0.4367 
          & 19.16 / 0.4929 & \underline{21.13} / \underline{0.5510} \\
    \midrule

\multirow{9}{*}{\shortstack{Circular $\frac{2r}{H}=1.0$\\ transparency rate $\approx 0.796$}} 
    & \multirow{3}{*}{15}
        & 1 & 9.26 / 0.0821 & 14.52 / 0.3324 & \textbf{21.19} / \textbf{0.6161} 
          & 12.50 / 0.3211 & 14.05 / \underline{0.5113} 
          & 17.94 / 0.2827 & \underline{19.85} / 0.4820 \\
    &   & 2 & 10.91 / 0.1247 & 16.54 / \underline{0.5477} & \textbf{22.06} / \textbf{0.6783} 
          & 16.19 / 0.4880 & 15.64 / 0.5357 
          & 19.27 / 0.4473 & \underline{21.60} / 0.5364 \\
    &   & 4 & 12.74 / 0.1840 & 17.15 / \underline{0.6457} & \textbf{23.78} / \textbf{0.7036} 
          & 16.76 / 0.5442 & 16.36 / 0.5424 
          & 19.60 / 0.5285 & \underline{22.73} / 0.5902 \\
    \cmidrule(l){2-10}

    & \multirow{3}{*}{25}
        & 1 & 9.25 / 0.0807 & 14.74 / 0.3525 & \textbf{21.12} / \textbf{0.6133} 
          & 13.76 / 0.3409 & 14.28 / \underline{0.5180} 
          & 18.23 / 0.3157 & \underline{19.82} / 0.4805 \\
    &   & 2 & 10.91 / 0.1227 & 16.55 / \underline{0.5593} & \textbf{21.90} / \textbf{0.6740} 
          & 16.31 / 0.4936 & 15.73 / 0.5364 
          & 19.25 / 0.4560 & \underline{21.55} / 0.5338 \\
    &   & 4 & 12.78 / 0.1815 & 17.09 / \underline{0.6462} & \textbf{23.70} / \textbf{0.7011} 
          & 16.69 / 0.5424 & 16.48 / 0.5435 
          & 19.55 / 0.5281 & \underline{22.70} / 0.5898 \\
    \cmidrule(l){2-10}

    & \multirow{3}{*}{50}
        & 1 & 9.23 / 0.0741 & 15.33 / 0.4935 & \textbf{20.76} / \textbf{0.5997} 
          & 15.80 / 0.4589 & 14.54 / \underline{0.5196} 
          & 18.72 / 0.4424 & \underline{19.65} / 0.4763 \\
    &   & 2 & 10.92 / 0.1137 & 16.37 / \underline{0.5949} & 21.23 / \textbf{0.6578} 
          & 16.24 / 0.5160 & 16.17 / 0.5378 
          & 19.12 / 0.5032 & \textbf{21.36} / 0.5260 \\
    &   & 4 & 12.87 / 0.1697 & 17.16 / 0.5494 & \textbf{23.38} / \textbf{0.6880} 
          & 16.70 / 0.4218 & 17.03 / 0.5450 
          & 19.33 / 0.5329 & \underline{22.54} / \underline{0.5828} \\
\midrule

\multirow{9}{*}{\shortstack{Annular\\ transparency rate $\approx 0.704$}}
    & \multirow{3}{*}{15}
        & 1 & 9.11 / 0.0786 & 12.97 / 0.2993 & \textbf{20.68} / \textbf{0.5996} 
          & 11.37 / 0.1431 & 13.26 / \underline{0.4882} 
          & 17.67 / 0.2550 & \underline{19.64} / 0.4741 \\
    &   & 2 & 10.61 / 0.1205 & 14.28 / 0.4962 & \textbf{21.09} / \textbf{0.6103} 
          & 13.90 / 0.4538 & 14.20 / 0.5028 
          & 19.15 / 0.4306 & \underline{21.30} / \underline{0.5240} \\
    &   & 4 & 12.13 / 0.1789 & 14.65 / \underline{0.5982} & \underline{21.40} / \textbf{0.6196} 
          & 14.39 / 0.5169 & 14.63 / 0.5086 
          & 19.49 / 0.5144 & \textbf{22.44} / 0.5830 \\
    \cmidrule(l){2-10}

    & \multirow{3}{*}{25}
        & 1 & 9.12 / 0.0773 & 13.09 / 0.3136 & \textbf{20.64} / \textbf{0.5974} 
          & 12.06 / 0.3058 & 13.28 / \underline{0.4888} 
          & 18.00 / 0.2850 & \underline{19.61} / 0.4742 \\
    &   & 2 & 10.63 / 0.1187 & 14.25 / 0.5087 & \underline{21.01} / \textbf{0.6072} 
          & 14.00 / 0.4599 & 14.22 / 0.5033 
          & 19.17 / 0.4410 & \textbf{21.24} / \underline{0.5225} \\
    &   & 4 & 12.18 / 0.1767 & 14.58 / \underline{0.5998} & \underline{21.32} / \textbf{0.6170} 
          & 14.31 / 0.5143 & 14.70 / 0.5092 
          & 19.45 / 0.5149 & \textbf{22.40} / 0.5821 \\
    \cmidrule(l){2-10}

    & \multirow{3}{*}{50}
        & 1 & 9.12 / 0.0714 & 13.34 / 0.4385 & \textbf{20.28} / \textbf{0.5849} 
          & 13.54 / 0.4165 & 13.45 / \underline{0.4907} 
          & 18.64 / 0.4281 & \underline{19.44} / 0.4692 \\
    &   & 2 & 10.70 / 0.1107 & 13.99 / \underline{0.5463} & \underline{20.69} / \textbf{0.5953} 
          & 13.85 / 0.4832 & 14.52 / 0.5052 
          & 19.08 / 0.4873 & \textbf{21.04} / 0.5149 \\
    &   & 4 & 12.38 / 0.1663 & 14.61 / 0.3617 & \underline{20.96} / \textbf{0.6044} 
          & 14.38 / 0.3251 & 15.09 / 0.5120 
          & 19.30 / 0.5178 & \textbf{22.20} / \underline{0.5746} \\

    \bottomrule
\end{tabular}
}
\label{tab:main_results}
\end{center}
\end{table*}

In this section, we evaluate the proposed PGD-MC algorithm and compare its performance with a commonly used cropping-based baseline and the CPnP-EM method \cite{pellizzari2020coherent}. 
We first consider a cropping-based approach in which a square sub-area of the hologram spectrum is extracted and the aperture is ignored during reconstruction \cite{pellizzari2017demonstration}. 
This strategy avoids the ill-posed forward operator by setting $A=I$, at the cost of reduced spatial resolution and sub-optimal reconstruction quality.
We then compare PGD-MC with the CPnP-EM method, which is  an alternating direction method of multipliers (ADMM) PnP based framework, termed as Multi-Agent Consensus Equilibrium \cite{buzzard2018plug}. CPnP-EM addresses the full forward model but avoids evaluating the exact likelihood gradient by maximizing a variational lower bound and further simplifying the updates using the approximation $A^H A \approx I$.
For completeness and to ensure a consistent implementation, we re-derive the CPnP-EM updates used in our experiments in Appendix~\ref{sec:app_B}. 

Both PGD-MC and CPnP-EM are model-based iterative reconstruction frameworks and can incorporate different priors through denoising or projection step.
To study the impact of the prior model, we consider three representative regularizers: classical BM3D denoiser \cite{dabov2007image}, pre-trained DnCNN denoiser \cite{zhang2017beyond}, and the untrained Deep Decoder \cite{heckel2018deep}.
We report PSNR and SSIM \cite{wang2004image} averaged over seven test images, provide representative visual comparisons, and analyze the computational behavior of PGD-MC in terms of Monte Carlo sampling and conjugate gradient convergence.

\subsection{Experimental setup}

We evaluate reconstruction performance using seven standard grayscale images of size $256 \times 256$ (\textit{barbara, boats, foreman, house, monarch, parrots, peppers}). 
Measurements are generated according to the multi-look forward model in \eqref{eq:forward_1}.
We consider different additive noise levels $\sigma_z$, numbers of looks $L$, and aperture configurations.
Specifically, we evaluate circular and annular apertures, with circular radii $r$ such that 
$\frac{2r}{H} \in \{0.8, 1.0\}$ corresponding to the transparency rates in Table~\ref{tab:main_results}. All simulations were run on NVIDIA RTX 6000 Ada GPUs.

All algorithmic hyperparameters and implementation details for PGD-MC and CPnP-EM are summarized in Table~\ref{tab:hyperparam}.

\begin{table}[t]
\centering
\caption{Summary of PGD-MC and CPnP-EM Parameter Settings}
\begin{tabular}{l c}
\toprule
\textbf{PGD-MC component} & \textbf{Setting} \\
\midrule
PGD iterations & 150 total, initialization $\frac{1}{L} \sum^L_{\ell=1}|A^H \yv_\ell|^2$ \\
Step size $\mu$ & 0.01 ($128 \times 128, 256\times256$),\; 0.005 ($512\times512$) \\
\midrule
Deep Decoder structure & Refer to \cite{heckel2018deep} \\
Kernel size & 1 ($128 \times 128, 256\times256$),\; 3 ($512\times512$) \\
Decoder channels & $[100,\;50,\;25,\;10]$ \\
Deep Decoder iters & 200 ($L{=}1$),\; 600 ($L{=}2$),\; 1000 ($L{=}4$) \\
Optimizer & Adam, learning rate $=1 \times 10^{-3}$ \\ \midrule 
BM3D noise level  & 25 ($L{=}1$),\; 25 ($L{=}2$),\; 25 ($L{=}4$)\\ \midrule
DnCNN noise level  & 50 ($L{=}1$),\; 50 ($L{=}2$),\; 50 ($L{=}4$)\\
\specialrule{0.8pt}{0pt}{0pt} 
\textbf{CPnP-EM component} & \textbf{Setting} \\
\midrule
EM iterations & 50 total, initialization $\frac{1}{L} \sum^L_{\ell=1}|A^H \yv_\ell|^2$ \\
Proximal strength $\sigma$ & $0.1$ \\
Convergence rate $\rho$ for Mann iteration & $0.2$ \\
\midrule
BM3D noise level  & 100 ($L{=}1,L{=}2,L{=}4$)\\ \midrule
DnCNN noise level  & 125 ($L{=}1,L{=}2,L{=}4$)\\
\bottomrule
\end{tabular}

\label{tab:hyperparam}
\end{table}

\subsection{Sub-area of hologram spectrum for reconstruction}

A common approach in digital holography is to sidestep the ill-posed aperture-induced forward model by reconstructing only a square sub-area of the hologram spectrum \cite{pellizzari2017demonstration}.
Cropping the spectrum effectively removes the aperture effect, reducing the forward operator to the identity $A = I$. Under this simplified model, reconstruction becomes a despeckling problem with diagonal covariance $\Sigma(\xv) = X + \sigma_z^2 I_n$, allowing the exact likelihood gradient to be computed efficiently.

While this approach is computationally efficient and numerically stable, it comes at the cost of reduced spatial resolution and sub-optimal reconstruction quality, since high-frequency components outside the cropped region are discarded.
The resolution of the reconstructed image is determined by the size of the cropped sub-area.
In our experiments, we reconstruct images of sizes $H' \times H'$ with $H' \in \{128,160\}$ from the original $H \times H$ hologram.
To enable fair comparison with full-resolution reconstructions, the recovered intensity is rescaled by a factor of $H'/H$, which compensates for the change in Fourier-domain sampling density introduced by cropping the 2D Fourier transform from size $H \times H$ to $H' \times H'$.

Figure~\ref{fig:subarea_psnr_convergence}(a) shows the reconstruction performance of the cropping-based baseline.
While the simplified model enables exact likelihood optimization, the reconstructions are resolution-limited and consistently underperform methods that account for the full aperture and ill-posed forward operator. This comparison highlights the need for approaches such as PGD-MC and CPnP-EM that directly handle the ill-posed model without sacrificing spatial resolution.

\begin{figure*}[h]
    \centering
    {\footnotesize
    \begin{subfigure}{0.48\textwidth}
        \centering
        \includegraphics[width=\linewidth]{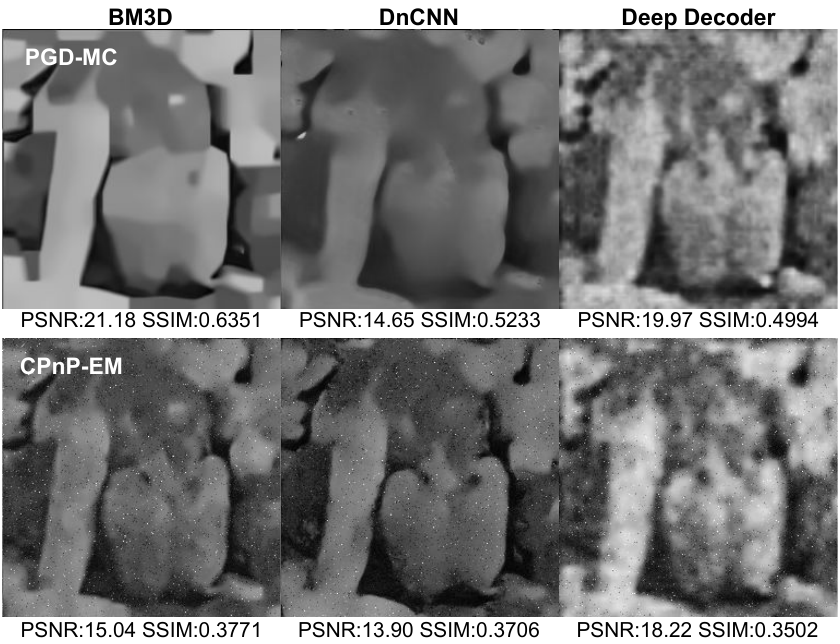}
        \caption{}
        \label{fig:vis_prior}
    \end{subfigure}
    \hfill
    \begin{subfigure}{0.48\textwidth}
        \centering
        \includegraphics[width=\linewidth]{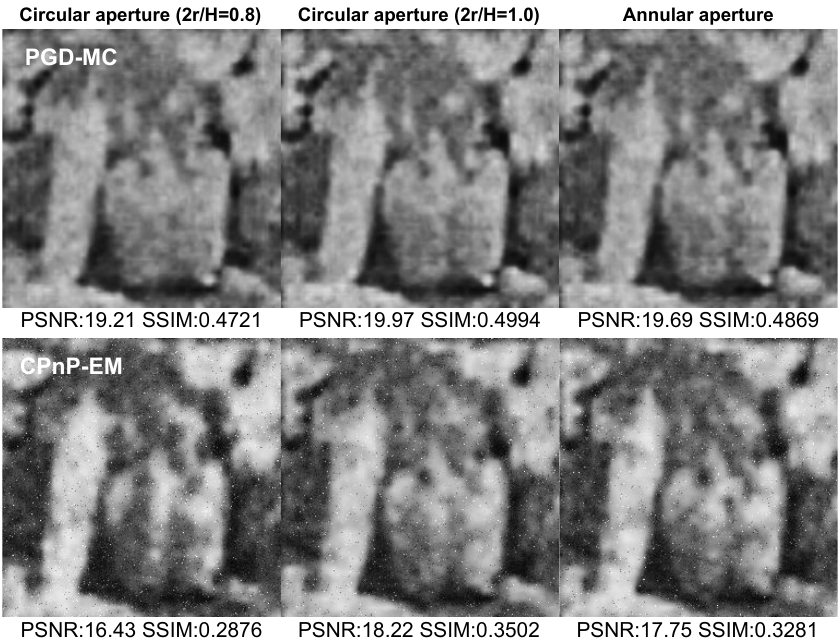}
        \caption{}
        \label{fig:vis_aperture}
    \end{subfigure}

    \vspace{0.6em}

    \begin{subfigure}{0.48\textwidth}
        \centering
        \includegraphics[width=\linewidth]{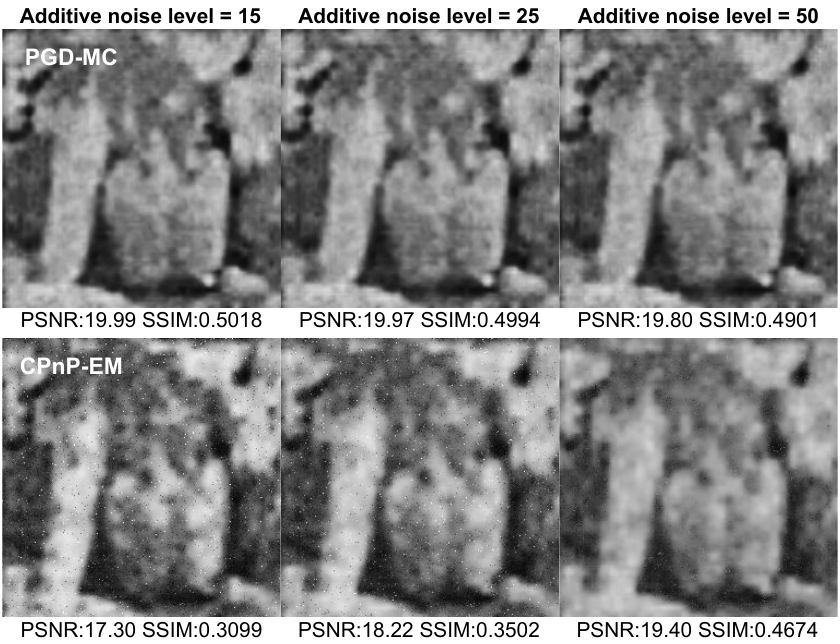}
        \caption{}
        \label{fig:vis_noise}
    \end{subfigure}
    \hfill
    \begin{subfigure}{0.48\textwidth}
        \centering
        \includegraphics[width=\linewidth]{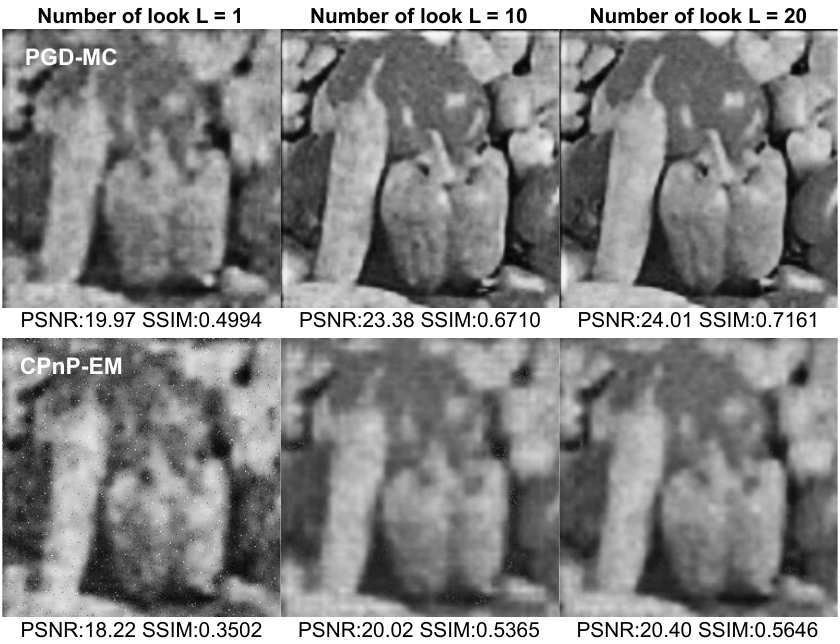}
        \caption{}
        \label{fig:vis_looks}
    \end{subfigure}
    }

    \caption{
        Reconstructions of PGD-MC (\textbf{top}) and CPnP-EM (\textbf{bottom}) under different experimental settings. In (b)-(d) Deep Decoder is used to model signal prior.
        (\textbf{a}) Different prior models, circular aperture radius $\frac{2r}{H} = 1.0$, single look, and additive noise $\sigma_z=25$. 
        (\textbf{b}) Different aperture models (circular and annular), single look, additive noise $\sigma_z=25$. 
        (\textbf{c}) Different noise levels $\sigma_z=15,25,50$, circular aperture radius $\frac{2r}{H}=1.0$, single look. 
        (\textbf{d}) Different numbers of looks $L=1,10,20$, circular aperture radius $\frac{2r}{H}=1.0$, additive noise $\sigma_z=25$.
    }
    \label{fig:combined_prior_aperture_noise_looks}
\end{figure*}

\subsection{Reconstruction performance}

In this section, we evaluate PGD-MC and CPnP-EM using the same set of prior models and experimental settings, where full forward model is considered.
Specifically, both methods are tested with BM3D, DnCNN, and Deep Decoder regularization across circular and annular apertures, additive noise levels $\sigma_z \in \{15,25,50\}$, and numbers of looks $L \in \{1,2,4\}$.
Table~\ref{tab:main_results} reports the average PSNR and SSIM over seven test images.

As illustrated by the PSNR convergence curves in Fig.~\ref{fig:subarea_psnr_convergence} (b), CPnP-EM exhibits unstable convergence behavior, particularly for small $L$ and low $\sigma_z$.
To enable a fair comparison, we therefore report the best PSNR achieved over the iterations for CPnP-EM.
In contrast, PGD-MC demonstrates stable convergence across all tested settings.

Overall, PGD-MC consistently outperforms CPnP-EM in terms of PSNR and SSIM across all experimental configurations, with particularly strong performance when combined with BM3D or Deep Decoder priors.
We next analyze the impact of individual factors, including the choice of prior model, aperture configuration, noise level, and number of looks.

\subsubsection{Impact of prior model}

We first examine the impact of the prior model on reconstruction performance.
Among the three evaluated priors, the Deep Decoder generally provides the best overall performance for both PGD-MC and CPnP-EM, in terms of both PSNR/SSIM and visual quality, as illustrated in Fig.~\ref{fig:vis_prior}.
In particular, Deep Decoder reconstructions preserve fine image structures while effectively suppressing speckle artifacts. When combined with PGD-MC, BM3D achieves competitive or higher PSNR values in several settings.
However, as shown in Fig.~\ref{fig:vis_prior}, BM3D-based reconstructions tend to exhibit noticeable over-smoothing, resulting in loss of fine details.
DnCNN shows comparable behavior but is more sensitive to the choice of denoising strength associated with the noise level of data the model is trained on.

From a practical perspective, Deep Decoder is also less sensitive to hyperparameter selection, since it does not require explicit specification of a noise level.
In contrast, the performance of BM3D and DnCNN depends strongly on the chosen noise parameter, and selecting an appropriate value in the presence of both speckle and model mismatch is nontrivial in practice.

\begin{table}[t] 
\caption{Reconstruction performance when $\sigma_z=0$ in measurement acquisition, and set $\sigma'_z=15,25,50$ in the reconstruction algorithms. The circular aperture used has radius $2r/H=1.0$.} 
\begin{center}

\begin{tabular}{ccccc}
    \toprule

    Real $\sigma_z$ & Hyperparam $\sigma'_z$ & Looks $L$ 
    & CPnP-EM & PGD-MC \\
    \midrule

    \multirow{9}{*}{0}
    & \multirow{3}{*}{15}
       & 1  & 17.94 / 0.2811 & 19.82 / 0.4827  \\
    &   & 2  & 19.28 / 0.4479 & 21.52 / 0.5341 \\
    &   & 4  & 19.63 / 0.5289 & 22.67 / 0.5899 \\
    \cmidrule(l){2-5}

    & \multirow{3}{*}{25}
       & 1  & 18.24 / 0.3149 & 19.72 / 0.4818 \\
    &   & 2  & 19.29 / 0.4397 & 21.37 / 0.5333 \\
    &   & 4  & 19.62 / 0.5291 & 22.49 / 0.5883 \\
    \cmidrule(l){2-5}

    & \multirow{3}{*}{50}
       & 1  & 18.90 / 0.4325 & 19.26 / 0.4779 \\
    &   & 2  & 19.32 / 0.4973 & 20.19 / 0.5255 \\
    &   & 4  & 19.57 / 0.5276 & 21.16 / 0.5826 \\
    \bottomrule
\end{tabular}
\label{tab:additive_results}
\end{center}
\end{table}

\subsubsection{Impact of aperture models}

We next evaluate the robustness of the reconstruction algorithms under different aperture configurations, including circular and annular apertures with varying transparency.
As shown in Fig.~\ref{fig:vis_aperture}, PGD-MC exhibits relatively stable performance across different aperture types and transparency levels.
This behavior is expected, since the proposed method explicitly incorporates the exact forward operator associated with each aperture into the likelihood model and gradient computation.
As a result, reconstruction quality improves with increasing aperture transparency, consistent with the increased amount of captured spatial-frequency information.

In contrast, the performance of CPnP-EM degrades more noticeably as aperture transparency decreases.
This sensitivity can be attributed to the approximation $A^H A \approx I$ used in CPnP-EM, which is exact only when the aperture has full transparency.
As the aperture becomes more restrictive, this approximation becomes increasingly inaccurate, leading to larger model mismatch and reduced reconstruction quality, as observed in Fig.~\ref{fig:vis_aperture}.

\subsubsection{Impact of  additive noise power}

We evaluate the sensitivity of the reconstruction algorithms to the additive noise level $\sigma_z$ in the measurement model.
Figure~\ref{fig:vis_noise} shows representative reconstructions obtained under different noise levels.
PGD-MC exhibits stable performance as $\sigma_z$ varies, with reconstruction quality improving as the additive noise level decreases.

In contrast, CPnP-EM is more sensitive to $\sigma_z$.
As detailed in Appendix~\ref{sec:app_B}, the E-step of CPnP-EM involves the posterior mean and covariance of $\gv$, whose numerical stability depends on the assumed additive noise level.
When $\sigma_z$ is small, the updates become ill-conditioned, leading to unstable convergence and degraded reconstruction quality, as observed in Fig.~\ref{fig:vis_noise}.

In all experiments above, we set the noise parameter used in reconstruction to match the true noise level in the measurements.
In practice, however, the assumed noise level $\sigma'_z$ in the reconstruction algorithm may differ from the true $\sigma_z$.
This situation is particularly relevant when speckle noise dominates and the additive noise component is negligible.
Setting $\sigma'_z = 0$ is undesirable, as the covariance matrix becomes $AXA^H$ which is singular; therefore, a small positive $\sigma'_z$ is required for numerical stability.

To study the effect of noise mismatch, Table~\ref{tab:additive_results} reports reconstruction performance when the true measurement noise is $\sigma_z = 0$, while the reconstruction algorithm assumes $\sigma'_z \in \{15,25,50\}$.
We observe that increasing $\sigma'_z$ can improve the performance of CPnP-EM when the number of looks is small.
However, as the number of looks increases (e.g., $L=4$), larger mismatch between $\sigma_z$ and $\sigma'_z$ leads to performance degradation.

\begin{table}[t] 
\caption{Reconstruction performance (Average PSNR and SSIM) when imaging model is $\yv = A\sqrt{\xv} + \zv$, with different $\sigma_z$, circular apertures with different $r$, PGD with BM3D is adopted to solve \eqref{eq:MLE_deblurring}.} 
\begin{center}
\begin{tabular}{ccc}
    \toprule
    Circular aperture & Additive noise level $\sigma_z$ 
    & PGD with BM3D \\
    \midrule
    \multirow{3}{*}{$\frac{2r}{H}=0.8$}
        & 15  & 30.60 / 0.8838  \\
       & 25  & 28.74 / 0.8389  \\
       & 50  & 25.57 / 0.7410  \\
    \cmidrule(l){1-3}

    \multirow{3}{*}{$\frac{2r}{H}=1.0$}
       & 15  & 31.38 / 0.8907  \\
       & 25  & 29.17 / 0.8430  \\
       & 50  & 25.56 / 0.7439  \\
    \bottomrule
\end{tabular}
\label{tab:denoising_results}
\end{center}
\end{table}

\subsubsection{Impact of number of looks}

We evaluate the effect of the number of looks $L$ on reconstruction performance.
Increasing $L$ improves the quality of the initial estimate $\frac{1}{L} \sum_{\ell=1}^L |A^H \yv_\ell|^2$ and reduces variance in both the likelihood gradient in \eqref{eq:grad} and the posterior mean estimation of $\gv$ in CPnP-EM (Appendix~\ref{sec:app_B}).
As a result, multi-look acquisition is expected to improve reconstruction performance \cite{chen2025multilook}.

This behavior is confirmed empirically in Fig.~\ref{fig:vis_looks}, where increasing the number of looks consistently leads to improved reconstruction quality for both PGD-MC and CPnP-EM.

\subsection{Efficiency of gradient computation}

We next evaluate the computational efficiency of PGD-MC.
Since the likelihood gradient computation involves both Monte Carlo (MC) sampling and conjugate gradient (CG) iterations, the overall runtime depends primarily on the number of MC samples and the convergence behavior of the CG solver.

\begin{figure*}[t]
    \centering
    {\footnotesize
    \begin{subfigure}{0.33\textwidth}
        \centering
        \includegraphics[width=\linewidth]{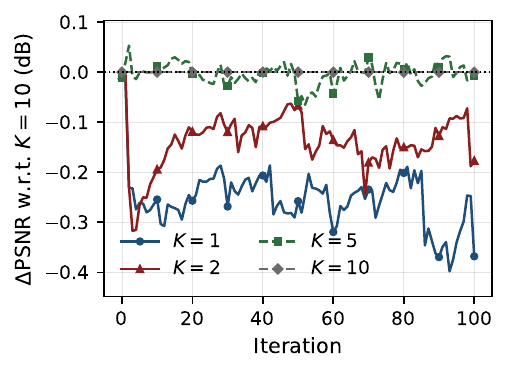}
    \end{subfigure}
    \hspace{-0.8em}
    \begin{subfigure}{0.33\textwidth}
        \centering
        \includegraphics[width=\linewidth]{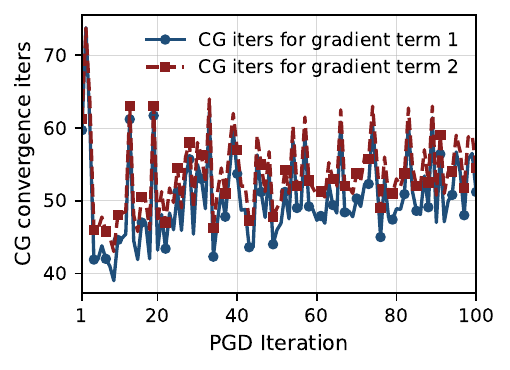}
    \end{subfigure}
    \hspace{-0.8em}
    \begin{subfigure}{0.335\textwidth}
        \centering
        \includegraphics[width=\linewidth]{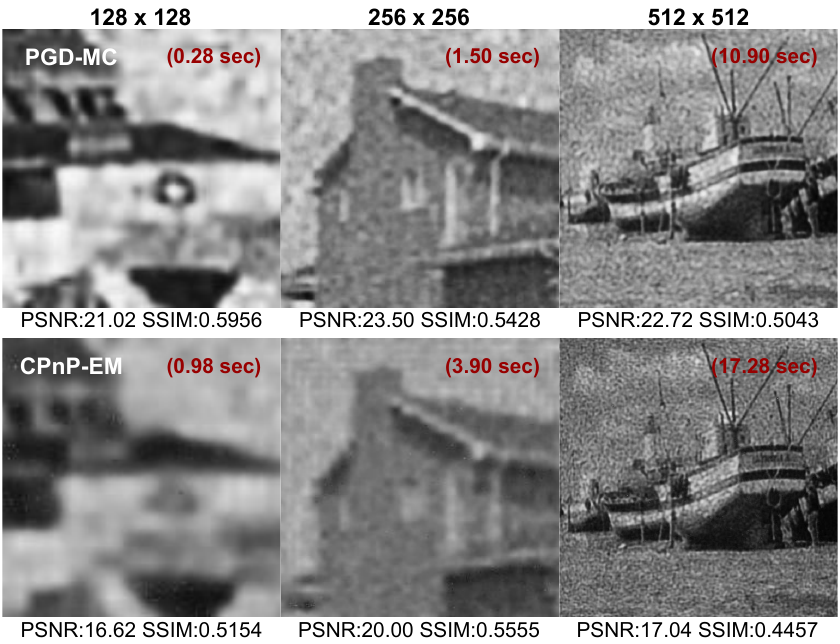}
    \end{subfigure}
    }

    \caption{
        Effect of MC samples $K$ on reconstruction performance (\textbf{left}). 
        Number of CG iterations required to converge at each PGD iteration (\textbf{middle}), test image is \textit{peppers}, aperture radius $\frac{2r}{H}=1.0$, single look $L=4$, 
        additive noise $\sigma_z = 25$. 
        Time cost per iteration for gradient approximation in PGD-MC and CPnP-EM (\textbf{right}), image sizes are 
        $128{\times}128$, $256{\times}256$, and $512{\times}512$, with $K=5$ MC samples.
    }
    \label{fig:MC_CG_time}
\end{figure*}

\paragraph{Monte Carlo sample size}
The MC approximation of the diagonal of $A^H \Sigma(\xv_t)^{-1} A$ requires a sufficient number of samples to achieve accurate gradient estimates.
We study the effect of the number of MC samples $K$ on reconstruction performance.
Figure~\ref{fig:MC_CG_time} (left) shows PSNR convergence curves for PGD-MC with $K \in \{1,5,10,20\}$, evaluated on the $256 \times 256$ image \textit{peppers}.
The results indicate that a small number of samples ($K=5$) is sufficient to achieve stable reconstruction performance, with diminishing returns for larger $K$.

\paragraph{Conjugate gradient convergence}
The CG solver is used to compute the matrix–vector products required in \eqref{eq:cg_1} and \eqref{eq:cg_2}.
We examine the number of CG iterations required to reach convergence at each PGD iteration, with the stopping tolerance in Algorithm~\ref{alg:CG} set to $\epsilon = 10^{-6}$.
Figure~\ref{fig:MC_CG_time} (middle) reports the number of CG iterations per PGD iteration for the $256 \times 256$ \textit{peppers} image.
Across iterations, CG typically converges within approximately 45--60 iterations, indicating stable and predictable inner-loop behavior.

\paragraph{Run-time analysis}
The computational cost of each PGD iteration is dominated by the CG solves required for gradient evaluation.
Specifically, computing the term $\diag(A^H \Sigma(\xv_t)^{-1} A)$ requires $K$ CG solves, while evaluating $\frac{1}{L} \sum_{\ell=1}^L A^H \Sigma(\xv_t)^{-1} \yv_\ell$ requires $L$ CG solves.
Consequently, the total runtime per PGD iteration can be approximated as
\[
(K+L) \times \text{(CG iterations to converge)} \times \text{(time per CG iteration)}.
\]

By exploiting the FFT-based structure of $A$ and $\Sigma(\xv)$, each CG iteration can be implemented efficiently.
In our implementation, the average time per CG iteration is approximately $1 \times 10^{-3}$, $5 \times 10^{-3}$, and $2 \times 10^{-2}$ seconds for image sizes of $128 \times 128$, $256 \times 256$, and $512 \times 512$, respectively.

For comparison, CPnP-EM performs per-coordinate root-finding in the M-step (Appendix~\ref{sec:app_B}, \eqref{eq:M_step_roots}) and requires additional computations in the E-step for each look.
Figure~\ref{fig:MC_CG_time} (right) reports the per-iteration computation time of PGD-MC and the combined E-step and M-step time of CPnP-EM (excluding denoiser application) for different image sizes.
PGD-MC achieves lower per-iteration runtime while providing higher reconstruction quality, and its computational advantage becomes more pronounced as image resolution increases.

Scalability to high-resolution imaging is particularly important in practical coherent imaging systems, where increasing sensor dimensions effectively enlarges the physical aperture and improves achievable spatial resolution \cite{tippie2011high, gao2022resolution}.

\subsection{Challenges in coherent imaging reconstruction}

To highlight the  impact of speckle noise and the difficulty of coherent imaging reconstruction in the presence of speckle noise,  consider $\yv = A \sqrt{\xv} + \zv$, a simplified speckle-free reference scenario. Here, $\zv$, as before, denotes the additive white Gaussian noise. The corresponding MLE optimization can be written as
\begin{align} \label{eq:MLE_deblurring}
    \hat{\xv} = \argmin_{\xv \in \Xc} \|\yv - A \sqrt{\xv}\|^2_2.
\end{align}
Table~\ref{tab:denoising_results} shows the reconstruction performance achieved by using PGD to solve  \eqref{eq:MLE_deblurring}.  It can be observed that compared to the speckle-corrupted measurements studied before, this simplified setting yields substantially higher PSNR and SSIM values. More importantly, unlike the case where speckle noise is present, in this simplified setting, changing the noise power ($\sigma_z$) considerably impacts the reconstruction performance.  This comparison further underscores the challenges associated with recovery in the presence of  multiplicative speckle noise.

\section{Conclusion}
We proposed a scalable, matrix-free optimization framework for maximum likelihood reconstruction in coherent imaging with speckle noise.
By combining randomized trace estimation with conjugate gradient methods, the proposed PGD-MC algorithm enables direct optimization of the exact likelihood without explicit covariance matrix formation or simplifying assumptions on the forward operator.
Experimental results on digital holography imaging model demonstrate improved reconstruction quality and stable convergence across a range of aperture configurations, noise levels, and numbers of looks.

\appendix


\section{MLE gradient derivation}\label{sec:app_A}

Starting from the negative log-likelihood defined in \eqref{eq:objective},
we derive the gradient expression in \eqref{eq:grad}.
Only intermediate steps omitted in the main text are provided.

Let $dX = \operatorname{diag}(d\xv)$ denote a perturbation of the reflectivity,
which induces the covariance perturbation
\[
d\Sigma = A\, dX\, A^H .
\]
Using standard matrix differential identities, the differential of $f_L(\xv)$
can be written as
\[
df_L
= \operatorname{tr}\!\big( \Sigma^{-1} d\Sigma \big)
- \frac{1}{L} \sum_{\ell=1}^L 
\yv_\ell^{H} \Sigma^{-1} d\Sigma \,\Sigma^{-1} \yv_\ell .
\]
Substituting $d\Sigma$ and using cyclic trace properties yields
\[
df_L
= \operatorname{tr}\!\big( A^H \Sigma^{-1} A \, dX \big)
- \frac{1}{L} \sum_{\ell=1}^L 
\operatorname{tr}\!\big( H_\ell \, dX \big),
\]
where $H_\ell =
\big(A^H \Sigma^{-1} \yv_\ell\big)
\big(A^H \Sigma^{-1} \yv_\ell\big)^H.$
Since $dX$ is diagonal, only the diagonal entries contribute, which leads
directly to the gradient expression in \eqref{eq:grad}.

\section{Derivation of CPnP-EM Updates}
\label{sec:app_B}
For completeness and to support reproducibility, we summarize the update equations of CPnP-EM \cite{pellizzari2020coherent}.
We emphasize the approximations used by CPnP-EM, since they directly affect robustness under non-ideal apertures and low-noise settings.
Following \cite{pellizzari2020coherent}, we re-derive the algorithm under an intensity-only speckle model, i.e., neglecting phase errors, consistent with the forward model in \eqref{eq:forward_1}.

\subsection{Surrogate objective and EM formulation.} We consider the coherent imaging model in \eqref{eq:forward_1}, where the speckle field $\gv$ is modeled as a latent Gaussian variable with prior $\mathcal{CN}(\mathbf{0},X)$.
 Instead of directly maximizing the likelihood $p(\yv \mid \xv)$, the CPnP-EM algorithm proposed in  \cite{pellizzari2020coherent} optimizes a variational lower bound of the marginal likelihood using an expectation–maximization procedure.

 For any auxiliary distribution $q(\gv)$ with support contained in that of $p(\yv,\gv \mid \xv)$, we have
\begin{align*}
    \log p(\yv \mid \xv) &= \log p(\yv, \gv \mid \xv)
    - \log p(\gv \mid \yv,\xv)\\
    &= \mathbb{E}_q \big[\log p(\yv, \gv \mid \xv)\big]
    - \mathbb{E}_q \big[\log p( \gv \mid \yv,\xv)\big] \\
    &=  \mathbb{E}_q \big[\log p(\yv,\gv \mid \xv)\big] + H(q) + D_{\mathrm{KL}}(q \| p).
\end{align*}
Define the negative lower bound as a  loss function defined on $(q, \xv)$ as
\[
\Lc(q,\xv)
= - \mathbb{E}_{q}[\log p(\yv, \gv \mid \xv)] - H(q).
\]
The EM algorithm is used in \cite{pellizzari2017demonstration,pellizzari2017phase,pellizzari2018optically,pellizzari2019imaging,pellizzari2020coherent} to optimize over $\xv, q$ to maximize the evidence lower bound (ELBO), $\Lc(q,\xv)$, of $\log p(\yv \mid \xv)$.

\subsection{Expectation–Maximization (EM) algorithm}
The EM algorithm alternates between maximizing the ELBO over $q$ (E-step)
and maximizing it over $\theta$ (M-step):
\begin{align*}
\text{E-step:} \quad &
q^{(t)}
=  \arg\max_{q}\Lc(q,\xv^{(t)})
= p(\gv \mid \yv,\xv^{(t)}),\\
\text{M-step:} \quad &
\xv^{(t+1)}
= \arg\max_{\xv}\Lc(q^{(t)},\xv).
\end{align*}
where $t$ denotes the iteration of EM algorithm.
\subsection{E-step: Posterior mean and covariance of $p(\gv|\yv, \xv)$}
In E-step, the variational distribution $q$ is set to $p(\cdot|\yv, \xv^{(t)})$. We know that $p(\cdot|\yv, \xv)$ is also complex Gaussian. Ignoring terms that are constant w.r.t $\gv$, the negative log-likelihood can be written as
\begin{align*}
-\log p(\yv, \gv \mid \xv)& = -\log p(\yv|\gv,\xv) - \log p(\gv|\xv)
\propto \frac{1}{\sigma_z^2}\|\yv-A \gv\|^2 + \gv^H X^{-1} \gv \\
&= \gv^H\!\Big(\tfrac{1}{\sigma_z^2}A^H A + X^{-1}\Big) \gv - 2\Re \Big\{ \big(\tfrac{1}{\sigma_z^2}A^H \yv \big)^H \gv\Big\}.
\end{align*}
Given $\yv$ and $\xv^{(t)}$, define $\bv  = \tfrac{1}{\sigma_z^2}A^H \yv$,
and
\begin{align*}
\mu^{(t)} = (\Lambda^{(t)})^{-1} \mathbf{b}, \quad \text{and}\; \Lambda^{(t)} = \tfrac{1}{\sigma_z^2}A^H A + (X^{(t)})^{-1}.
\end{align*}
Then, since $\gv^H\Lambda^{(t)} \gv - 2 \Re \{(\mathbf{b}^{(t)})^H \gv\}
\nonumber= (\gv-\mu^{(t)})^H \Lambda^{(t)} (\gv-\mu^{(t)}) - (\mu^{(t)})^H \Lambda^{(t)} \mu^{(t)}$,
$p(\gv|\yv,\xv^{(t)})=\mathcal{CN}(\mu^{(t)},C^{(t)})$, where
\[
C^{(t)}=(\Lambda^{(t)})^{-1}.
\]
Furthermore, it follows that
\[
\mathbb{E}[\gv\mid \xv^{(t)},\yv]=\mu^{(t)},\quad
\mathbb{E}[\gv\gv^H\mid \xv^{(t)},\yv]=C^{(t)}+\mu^{(t)}(\mu^{(t)})^H.
\]

In this formulation, $\Lambda^{(t)}$ and $C^{(t)}$ denote two   large $n\times n$ matrices. To avoid storing and inverting such  large  matrices, and derive an implementable version of the EM algorithm, typically $A^H A $ is approximated by a diagonal matrix as 
\[
A^H A \stackrel{\rm (a)}{=} \mathcal{F}^H M\mathcal{F} \approx I,
\]
where (a) follows because $M^HM=M$. The accuracy of this approximation depends on the aperture $M$ and how closeness to the identity map. For example, for the annular aperture  this approximation becomes  coarse and degrades the performance of the algorithm. 
Using this approximation, the diagonal entries of matrix $C^{(t)}$ and the entries of  vector $\mu^{(t)}$ can be written as 
\begin{align} \label{eq:E_step_posterior_mean}
    C^{(t)}_{ii} = \frac{\sigma^2_z \Bar{x}^{(t)}_i}{\sigma^2_z + \Bar{x}^{(t)}_i}, \text{and}\;
    \left|\mu^{(t)}_i\right|^2 = \frac{1}{\sigma^4_z} (C^{(t)}_{ii})^{2}\left|A^H\yv \right|_i^2,
\end{align}
respectively.

\subsection{M-step: Optimize the Q-function}
Next, $q^{(t)}(\mathbf{g})$ is fixed, and   $\mu^{(t)}$ and $C^{(t)}$ derived in the  E-step are used as constants. Hence, in the M-step we maximize $\Lc(\xv)$, where
$\Lc(\xv)= \mathbb{E}_{q^{(t)}(\gv)}[-\log p(\yv, \gv\mid \xv)] = \mathbb{E}_{p( \gv\mid \yv,\bar{\xv}^{(t)})}[-\log p(\yv, \gv\mid \xv)] = \mathbb{E}_{p(\gv\mid \yv,\bar{\xv}^{(t)})}\left[ -\log p(\yv|\gv) -\log p(\gv|\xv)\right] = \Lc(\xv; \bar{\xv}^{(t)}) + c$. 

This function captures the expected complete-data log-likelihood under the current posterior of the latent variable, where the loss terms w.r.t $\xv$ is
\begin{align*}
\Lc (\xv; \bar{\xv}^{(t)}) &= \mathbb{E}_{p(\gv\mid \yv,\bar{\xv}^{(t)})}\left[ -\log p(\gv|\xv)\right] \\ 
&=\mathbb{E}_{p(\gv\mid \yv, \bar{\xv}^{(t)})} \left[ \gv^H X^{-1} \gv \right] + \log \left| X \right|.
\end{align*}
Note that 
\begin{align*}
\mathbb{E}_{p(\gv\mid \yv,\bar{\xv}^{(t)})}\big[\gv^H X^{-1} \gv\big]= \mathrm{tr} \Big(X^{-1}\,\mathbb{E}[\gv\gv^H]\Big) = \mathrm{tr}\Big(X^{-1}\,(C+\mu\mu^H)\Big) = \sum_{i=1}^n \frac{C^{(t)}_{ii}+|\mu^{(t)}_i|^2}{x_i}.
\end{align*}
Thus the objective function simplifies to
\begin{align} \label{eq:surrogate_MLE}
\Lc(\xv; \xv^{(t)}) = \sum_{i=1}^n \frac{C^{(t)}_{ii}+|\mu^{(t)}_i|^2}{x_i} + \log x_i.
\end{align}

\subsection{M-step augmented by PnP  denoiser}
Authors in \cite{pellizzari2020coherent} use a Plug-and-Play approach to solve the M-step by incorporating a pre-trained DnCNN denoiser to enforce the source model through regularization. They define the following two optimizations
\begin{align}
&F_1(\xv^{(t)}_1; \Bar{\xv}^{(t)}) 
= \arg\min_{\xv} 
\left\{
  \Lc(\xv; \Bar{\xv}^{(t)}) 
  + \frac{1}{2\sigma^2} \| \xv - \xv^{(t)}_1 \|^2
\right\},  \notag \\
&F_2(\xv^{(t)}_2) 
= \arg\min_{\xv} 
\left\{
  \beta f_2(\xv) 
  + \frac{1}{2\sigma^2} \| \xv - \xv^{(t)}_2 \|^2
\right\}, \label{eq:M-step}
\end{align}
where $f_2(\xv) = - \log p(\xv)$. Here,  $\beta$ and $\sigma^2$ denote free parameters that need to be properly set.  Define 
\begin{align}
    \xv^{(t)} =
\begin{bmatrix}
    \xv^{(t)}_1\\
    \xv^{(t)}_2
\end{bmatrix},
\quad
\bar{\xv}^{(t)} =\frac{\xv^{(t)}_1 + \xv^{(t)}_2}{2}.
\end{align}
At the end of each M-step, $\bar{\xv}^{(t)}$ is going to be used for the next E-step. Furthermore, $  \xv^{(t)}$ is updated as follows:
\begin{align}
&\mathbf{w}^{(t+1)} \leftarrow
\begin{bmatrix}
F_1(\xv^{(t)}_1; \bar{\xv}^{(t)}) \\
F_2(\xv^{(t)}_2)
\end{bmatrix}, \\
&\xv^{(t+1)} \leftarrow 
\xv^{(t)} + 2\rho\big[ 
G(2\mathbf{w}^{(t+1)} - \xv^{(t)}) - \mathbf{w}^{(t+1)}
\big].
\end{align}
Here  $G$ is an \textit{averaging operator} defined as follows. Given $\xv =[\xv_1^T,\ldots,\xv_M^T]^T$, $G(\xv)=[\bar{\xv}^T,\ldots,\bar{\xv}^T]^T$, with $\bar{\xv}$ repeated $M$ times and 
$\bar{\xv} = \frac{1}{M} \sum^M_{i=1} \xv_i.$

The first optimization in \eqref{eq:M-step} is solved by taking the derivative of the objective function and setting it to zero. The objective function in $F_1$ is
\begin{align*}
\Lc_{F_1} &= \Lc(\xv; \Bar{\xv}^{(t)}) + \frac{1}{2\sigma^2} \| \xv - \xv^{(t)}_1 \|^2  \\ 
&= \sum_{i=1}^n \frac{C^{(t)}_{ii}+|\mu^{(t)}_i|^2}{x_i} + \log x_i + \frac{x^2_i - 2 x^{(t)}_{1,i} x_i + (x^{(t)}_{1,i})^{2}}{2 \sigma^2},
\end{align*}
for each coordinate $i$, the objective is
\begin{align} \label{eq:M_step_coordinate}
    \frac{C^{(t)}_{ii}+|\mu^{(t)}_i|^2}{x_i} + \log x_i + \frac{x^2_i - 2 x^{(t)}_{1,i} x_i + (x^{(t)}_{1,i})^{2}}{2 \sigma^2},
\end{align}
the gradient of it w.r.t $x_i$ is
\begin{align}
    \frac{\partial \Lc_{F_1}}{\partial x_i} = -(C^{(t)}_{ii}+|\mu^{(t)}_i|^2)\frac{1}{x^2_i} + \frac{1}{x_i} + \frac{1}{\sigma^2} (x_i - x^{(t)}_{1,i}).
\end{align}
To get minimizer $F_1$, we find the local minimum by getting roots when setting first-order derivative to 0,
\begin{align} \label{eq:M_step_roots}
    x^3_i - x^{(t)}_{1,i} x^2_i + \sigma^2 x_i - \sigma^2 (C^{(t)}_{ii}+|\mu^{(t)}_i|^2) = 0
\end{align}
solve this equation to obtain the roots, and select the positive real valued root which obtains lowest $\Lc_{F_1}$ as the minimizer $F_1$. This is performed on coordinates sequentially.

To solve the second optimization and find $F_2$, a pre-trained denoiser such as DnCNN \cite{zhang2017beyond}, is employed with a fixed noise level.

Finally, in multi-look setup \cite{bate2021model},   the E-step is modified as:
\begin{align} \label{eq:E_step_posterior_mean_multilook}
    \left|\mu^{(t)}_{L,i}\right|^2 = \frac{1}{\sigma^4_z} (C^{(t)}_{ii})^2 \frac{1}{L} \sum^L_{\ell=1} \left|A^H\yv_\ell\right|_i^2.
\end{align}

\begin{algorithm}[t]
  \caption{EM algorithm for C-PnP \cite{pellizzari2020coherent}}
  \label{alg:EM}
  \begin{algorithmic}[1]
    \STATE \textbf{Input:} Measurements $\{\yv_\ell\}_{\ell=1}^L$, forward operator $A$, denoiser $g_\theta$, proximal strength $\sigma$, convergence rate $\rho$ for Mann iteration, total number of iterations $T$.
    \STATE \textbf{Initialization:}
        $\bar{\xv}^{(1)} = \xv^{(1)}_1 = \xv^{(1)}_2
        = \frac{1}{L}\sum_{\ell=1}^L \bigl|A^{H}\yv_\ell\bigr|^2 .$
    \FOR{$t = 1,2,\ldots,T-1$}
      \STATE \textbf{[E-step]}
      \FOR{$i = 1,2,\ldots,n$}
        \STATE
          $C^{(t)}_{ii}
          = \frac{\sigma_z^2\,\bar{x}^{(t)}_i}{\sigma_z^2 + \bar{x}^{(t)}_i}, \:\:
          \bigl|\mu^{(t)}_i\bigr|^2
          = \frac{\bigl(C^{(t)}_{ii}\bigr)^2}{\sigma_z^4}
            \bigl| (A^{H}\yv)_i \bigr|^2 .$
      \ENDFOR
      \STATE Define
        $\mathcal{L}(\xv;\bar{\xv}^{(t)})$ as in \eqref{eq:surrogate_MLE}.
      \STATE \textbf{[M-step]}
      \STATE \textbf{Proximal update of $\xv^{(t)}_1$:}
      \STATE Solve
        $\wv^{(t+1)}_1
        = \arg\min_{\xv}
        \left\{
          \mathcal{L}(\xv;\bar{\xv}^{(t)})
          + \frac{1}{2\sigma^2}\|\xv - \xv^{(t)}_1\|_2^2
        \right\}$ coordinate-wise:
      \FOR{$i = 1,2,\ldots,n$}
        \STATE $\mathbf{w}^{(t+1)}_{1,i}$ as
        real positive root of \eqref{eq:M_step_roots}
        which minimizes \eqref{eq:M_step_coordinate}.
      \ENDFOR
      \STATE \textbf{Denoising of $\xv^{(t)}_2$:}\\
      $\mathbf{w}^{(t+1)}_2 \gets g_\theta\!\left(\xv^{(t)}_2\right)$.
      \STATE \textbf{Consensus update:}
      \STATE Form the stacked variable
    $\mathbf{w}^{(t+1)} \gets
        \begin{bmatrix}
          \mathbf{w}^{(t+1)}_1 \\
          \mathbf{w}^{(t+1)}_2
        \end{bmatrix}.$
      \STATE
        $\xv^{(t+1)} \gets
        \xv^{(t)}
        + 2\rho \Big(
          G\big(2\mathbf{w}^{(t+1)} - \xv^{(t)}\big)
          - \mathbf{w}^{(t+1)}
        \Big).$
      \STATE Split
        $\begin{bmatrix}
          \xv^{(t+1)}_1 \\
          \xv^{(t+1)}_2
        \end{bmatrix}
        \gets \xv^{(t+1)}$, compute $
        \bar{\xv}^{(t+1)}
        \gets \frac{\xv^{(t+1)}_1 + \xv^{(t+1)}_2}{2}$ for E-step usage in the next iteration.
    \ENDFOR
    \STATE \textbf{Output:} $\hat{\xv} = \bar{\xv}^{(T)}$.
  \end{algorithmic}
\end{algorithm}

\newpage
\section*{Acknowledgment}
\noindent X.C., A.M., S.J. were supported in part by ONR award no. N00014-23-1-2371. S.J. was supported in part by NSF CCF-2237538.

\newpage
\bibliographystyle{IEEEtran}
\bibliography{refs}
\end{document}